\documentclass[10pt, letterpaper, twocolumn]{article}

\usepackage{amsmath, amssymb, amsfonts, mathtools}

\usepackage{newtxtext, newtxmath}

\usepackage[left=0.5in, right=0.5in, top=0.5in, bottom=0.5in,
            columnsep=0.15in]{geometry}

\usepackage{graphicx}
\usepackage{booktabs}
\usepackage{longtable}
\usepackage{array}
\usepackage{tabularx}
\usepackage{caption}
\usepackage{microtype}

\usepackage[super, sort&compress]{natbib}
\usepackage[colorlinks=true, linkcolor=blue, citecolor=blue,
            urlcolor=blue]{hyperref}

\graphicspath{{figures/}}

\usepackage{titlesec}
\titleformat{\section}{\fontsize{13}{15}\bfseries}{\thesection}{0em}{}
\titleformat{\subsection}{\fontsize{11}{13}\bfseries}{\thesubsection}{0em}{}

\usepackage{fancyhdr}
\usepackage{xcolor}

\begin{document}

\twocolumn[{%
  \begin{center}
    {\Large\bfseries
     Fundamental Dynamical Units for Physics-Informed Structural Inference\\
     from Perturbation Time-Series in Networked Systems}

    \vspace{1em}
    Nima Nouri

    \vspace{0.5em}
    \textit{Artificial Intelligence for Science Innovation,
            AstraZeneca, Waltham, MA, USA}

    \vspace{0.3em}
    {\small Corresponding author: nima.nouri@astrazeneca.com}
  \end{center}

  \vspace{0.1em}
\section*{Abstract}

In networked dynamical systems, the parameter of primary mechanistic interest is signed
interaction structure.
Recovering this structure from perturbation time-series data is a fundamental identification
problem, compounded by three coupled obstacles: the combinatorial complexity of interaction
architectures, ambiguity of causal attribution under limited interventions, and
state-dependent dynamics that confound structural inference.
Each obstacle is structural in origin and calls for a structural solution.
We address these challenges by adopting a reductionist approach, introducing Fundamental
Dynamical Units (FDUs): signed three-node interaction patterns as composable primitives
that convert the interaction hypothesis space into a finite, constructive, and tractable
representation.
We show that local interaction structure determines the perturbation conditions required
to disentangle direct from relayed influence, making intervention design a structural
consequence of the FDU representation.
We embed FDU-regularized structural inference within a physics-informed neural ordinary
differential equation (ODE) whose
governing-equation constraint transforms structural hypotheses into verifiable dynamical
predictions, enabling joint recovery of interaction structure and perturbation-resolved
trajectories.
Validated on synthetic benchmarks with known ground truth, the framework supports
structural commitment, expressed through FDU primitives, motif-prescribed intervention
design, and physics-informed learning, as a principled basis for mechanistically
interpretable inference in networked dynamical systems.

  \vspace{2em}
}]


\section{Introduction}

Recovering the interaction structure of a networked dynamical system from trajectory
observations is fundamentally non-injective: distinct causal architectures can produce
responses that are indistinguishable without carefully targeted interventions
\textbf{\cite{Pearl2009,Spirtes2000,Hauser2012}}, and more observations of the wrong kind
deepen this ambiguity rather than resolving it.
The objective, recovering which system components regulate which others with what
directionality and sign, must be pursued under three compounding constraints: measurements
are partial and temporally sparse \textbf{\cite{Aalto2020}}, interventions are limited in
scope and amplitude \textbf{\cite{Sarmah2022}}, and the governing dynamics are state-dependent, meaning the same structural interaction can produce qualitatively
different observable signatures at different operating conditions \textbf{\cite{Runge2023}}.
These are not merely statistical challenges; they are structural, and demand structural solutions.

Scaling data volume and model capacity, the dominant paradigm in contemporary machine
learning, cannot resolve this difficulty.
The challenge is fundamentally one of structural identification, not merely statistical
estimation: more interventional data of the same design cannot in general resolve the
ambiguity in the interaction hypothesis space \textbf{\cite{HeGeng2008}}, and a more
expressive model can fit multiple competing structures with comparable fidelity
\textbf{\cite{Gutenkunst2007}}.
Existing approaches partition around this impasse without resolving it: many methods that
prioritize flexible trajectory fitting provide limited causal or mechanistic semantics,
while approaches that enforce interpretable structure through tightly specified ODEs can
often be brittle under model misspecification and incomplete interventional coverage
\textbf{\cite{Polynikis2009}}.
Neither resolves the underlying issue: mechanistically interpretable inference requires
structural principles built into the model itself, not imposed post hoc on unconstrained
fits or recovered by scale alone, motivating a deliberate return to first-principles design
\textbf{\cite{Scholkopf2021,PetersBook2017}}.

We formalize this challenge as three coupled design constraints: interaction-structure space
(constraining combinatorial complexity while preserving mechanistic expressivity);
intervention-grounded interpretation (disambiguating direct from relayed influence using
structurally targeted interventions); and state-dependent nonlinear dynamics (learning
dynamics that generalize across operating regimes while maintaining mechanistic meaning).
Taken together, these requirements are mutually constraining: representation, causal
attribution, and dynamical learning must be treated as coupled design constraints rather
than independent modeling choices.

Interaction-structure space constitutes the first obstacle: the space of possible signed
interaction architectures is combinatorially vast \textbf{\cite{EdwardsGlass2000}},
requiring a hypothesis space grounded in the structural origin of the identification
difficulty rather than in computational convenience alone.
The key insight is that multi-path confounding has a minimal structural origin: a directed
pair $j \to i$ carries no pathway ambiguity by itself; a third node is the minimal addition
that creates a relay route $j \to k \to i$ in parallel, making the triad the fundamental
unit of multi-path confounding.
We therefore represent network architectures as compositions of Fundamental Dynamical Units
(FDUs): signed three-node interaction patterns, each capturing a distinct local pattern of
direct and relayed regulatory influence.
As reusable building blocks, FDUs provide a compositional basis for assembling
network-scale architectures from local three-node patterns
\textbf{\cite{Milo2002,Alon2007,ManganAlon2003,Benson2016}}, grounding the hypothesis space
in the level at which confounding originates.
Each FDU goes beyond pairwise connectivity, coupling nearby edge choices and making
direct-versus-relayed pathways explicit within each three-node subgraph: a shift from
surface to volumetric triangulation \textbf{(Fig.~\ref{fig:1}a,b)}, just as surface meshes
encode boundary topology whereas volumetric representations impose interior structural
constraints.
The resulting FDU-compositional prior makes the interaction hypothesis space substantially
more tractable and ensures every learned structure carries interpretable mechanistic meaning.

Intervention-grounded interpretation constitutes the second obstacle: even when a model
reproduces observed trajectories, the underlying mechanism can remain ambiguous.
Apparent influence between two nodes may reflect a direct interaction, a relayed pathway
through an intermediate, or a mixture of both, and these alternatives cannot generally be
resolved from purely observational data \textbf{\cite{Pearl2009,Eberhardt2007}}.
We therefore treat intervention design not as an independent modeling choice but as a
structural consequence: in this framework, once the FDU representation identifies which
parallel pathways are present, the perturbation conditions required to disentangle direct
from relayed influence are prescribed by the local interaction structure.
Concretely, interventions are treated not merely as additional training data but as
structurally targeted probes \textbf{\cite{Eberhardt2007,Hauser2012}}: in our framework,
the local FDU type specifies which perturbation combinations differentially excite parallel
pathways, enabling attribution of observed responses to specific mechanistic alternatives.
This framing elevates interpretability to a design objective: model structure and inference
are organized to produce interaction hypotheses that are intervention-consistent and
amenable to empirical refutation.

State-dependent nonlinear dynamics constitutes the third obstacle: the observable effect of a
structural interaction depends not only on whether the interaction exists, but on where in
state space the system currently operates.
An interaction that is structurally present and causally active can produce near-zero
observable signal when the system's state places it in an insensitive regime, a property of
nonlinear systems that we refer to as operating-point dependence \textbf{\cite{Kent2013}}.
A model fitted at a single operating point or steady state cannot in general distinguish a
genuinely absent edge from one that is present but locally insensitive
\textbf{\cite{Stepaniants2020}}, and the effective interaction strengths it recovers reflect
local operating conditions rather than invariant structural parameters.
Addressing this requires continuous-time learning that tracks the system through its full
trajectory, including transient phases where structural interactions are most discriminable
\textbf{\cite{Stepaniants2020}}, with sufficient mechanistic grounding to separate structural
parameters from state-dependent response amplitudes.

Prior work on directed structure recovery from time-series data has developed along two
broad traditions.
The first comprises data-driven methods that recover interaction structure under parsimony
constraints: these span regression-based and Granger-causal frameworks
\textbf{\cite{Granger1969,Huynh-Thu2018}}, sparse system identification
\textbf{\cite{Brunton2016}}, perturbation-response network reconstruction
\textbf{\cite{Timme2014}}, directed acyclic graph (DAG) learning from time-series under linear dynamics
\textbf{\cite{Pamfil2020}}, and latent graph inference from dynamics including graph-neural-ODE approaches that infer cyclic directed graphs but not signed edges
\textbf{\cite{Kipf2018,Huang2020graphode,Bhaskar2024}}.
The second comprises causal formalizations that establish identifiability conditions for
structural recovery under dynamical interventions
\textbf{\cite{Peters2013timeseries,Runge2019}}, clarifying when causal
structure can in principle be distinguished from trajectory data under specific model
assumptions.
Both traditions make genuine progress on scalable recovery and causal semantics, but treat
the perturbation panel as a fixed input rather than as a structural consequence of local
interaction architecture, and do not organize the hypothesis space around local
direct-versus-relay ambiguity as the primary structural unit; neither addresses
operating-point dependence, leaving recovered interaction estimates conflated with
state-dependent response sensitivity rather than separated into invariant structural
parameters and operating-point-specific signal amplitudes.
A related body of work identifies network motifs as a post-hoc structural vocabulary
\textbf{\cite{Milo2002,Alon2007,ManganAlon2003}}, applying motif descriptions to networks
after structure has been inferred, with edge sign treated as a secondary attribute where
considered at all; the present framework instead uses an exhaustive signed encoding of
three-node patterns as the hypothesis space itself, making signed interaction type a
primary inferential target and motif structure the basis of inference rather than a
post-hoc descriptor.
This distinction extends to the nature of the inference output.
When the motif representation constitutes the hypothesis space, structural attribution,
identifying which signed primitives generate each recovered interaction and at what
compositional weight, is produced intrinsically by inference rather than by a secondary
analysis decoupled from the evidence.
In networks with multiple overlapping subgraphs, post-hoc assignment is ambiguous where
intrinsic attribution is not.

Building on this perspective, we turn to continuous-time neural dynamical models as a
substrate for hybrid inference.
Neural ordinary differential equations (NODEs) \textbf{\cite{Chen2018}} recast deep
networks as continuous-time dynamical systems by learning a vector field and integrating it
with differentiable solvers, yielding continuous trajectories and enabling end-to-end
training.
In practice, however, standard NODE formulations do not constrain the learned vector field
to correspond to a mechanistically interpretable interaction structure: different vector
fields can reproduce similar trajectories without yielding a unique, interpretable
decomposition into directed interactions.
Neural controlled differential equations (Neural CDEs) \textbf{\cite{Kidger2020}} address
part of this gap by driving latent dynamics with an explicit control path, improving the
treatment of irregular sampling and partial observability, yet the governing field is still
rarely mechanistically constrained.
In parallel, physics-informed approaches encourage dynamical consistency by penalizing
violations of differential-equation constraints during training, often improving plausibility
and temporal generalization \textbf{\cite{Raissi2019,Karniadakis2021,YuWang2024}};
nevertheless, such constraints alone do not specify which interaction structures generate
the observed dynamics, nor do they resolve ambiguity when multiple mechanisms are compatible
with the same trajectories.
Physics-informed approaches have been extended to inverse problems, recovering continuous
scalar or functional parameters from trajectory data, and sparse methods have identified
dominant dynamical terms without prescribed structure \textbf{\cite{Mircea2024,Brunton2016}}; neither
addresses the combinatorial problem of inferring signed directed topology, in which edge
existence, direction, and polarity must be jointly recovered from interventional
observations under relay ambiguity.
Taken together, continuous-time neural models tend to emphasize flexible data fit, whereas
physics-informed formulations tend to emphasize dynamical consistency; neither alone fully
satisfies the joint requirements of interpretable structure, intervention-grounded
attribution, and state-dependent mechanistic learning.
This motivates a physics-informed neural ODE perspective that couples continuous-time
dynamical learning with mechanistic constraints and an explicit representation of
interaction structure.

Accordingly, we develop a FDU-regularized physics-informed neural ODE framework that
makes multi-path confounding explicit and operational.
Specifically, this work contributes (i) an exhaustive, label-consistent encoding of signed
three-node interaction patterns into permutation-invariant structural classes; (ii) a
structure-derived perturbation design principle that grounds
panel selection in the identifiability requirements of local interaction architecture; and
(iii) a FDU-regularized physics-informed neural ODE that embeds this structured prior
within continuous-time dynamical learning for joint trajectory prediction and signed
interaction inference \textbf{(Fig.~\ref{fig:1}c,d)}.
We demonstrate the framework in a proof-of-concept demonstration on synthetic benchmarks with
known ground truth, establishing signed interaction recovery and motif-level attribution as
falsifiable expectations.
Together, these contributions establish a principled route toward compact, testable
hypotheses about multi-component dynamical systems from heterogeneous observational and
interventional time-series data.

\section{Results}

The following Results develop the structural theory from first principles; implementation
details are provided in Methods.

\subsection{The Signed Three-Node Interaction Space Admits an Exhaustive
and Constructive Dictionary of Fundamental Dynamical Units}

The signed three-node interaction space is the smallest structural level admitting both
direct and relay pathways between the same node pair, and therefore constitutes the natural
primitive space for motif-centric signed structural inference.
We sought to characterize this space exhaustively, enumerating all admissible signed
configurations and establishing a label-consistent encoding.

We consider signed, directed three-node systems in which each unordered pair $\{1,2\}$,
$\{2,3\}$, and $\{1,3\}$ is pairwise-connected, meaning that at least one directed influence
is present between the two nodes, the minimal condition for a relay route to coexist with a
direct edge between the same pair.
For each unordered pair of nodes $\{i,j\}$, the directed edges $i\to j$ and $j\to i$ can
each be absent, activating, or inhibiting, provided they are not simultaneously absent
(pairwise connectivity).
This yields eight admissible patterns per unordered pair: four single-direction signed
configurations (activating or inhibiting, in either direction) and four mutual configurations
(mutual activating, mutual inhibiting, and the two mixed-sign cases).
Because the three unordered pairs can be specified independently, the total number of
pairwise-connected signed triads is $8^3 = 512$.

We represent each signed triad by a ternary matrix $T\in\{0,1,2\}^{3\times3}$, with rows
indexing target nodes and columns indexing source nodes, so that $T_{ij}$ encodes the
influence of source $j$ on target $i$ (0: no edge; 1: activation $j\to i$; 2: inhibition
$j\dashv i$).
For example,
\begin{equation}
  T = \begin{pmatrix} 0 & 0 & 2 \\ 1 & 0 & 0 \\ 0 & 1 & 0 \end{pmatrix},
\end{equation}
encodes a negative-feedback cycle $1\to2\to3\dashv1$.
Flattening $T$ row-wise yields a 9-character motif code (here, \texttt{002100010}), which
serves as the canonical identifier for each triad in all subsequent figures and tables.

The triad encoding admits a compact, constructive dictionary of signed three-node primitives
for structural composition: the Fundamental Dynamical Unit (FDU) dictionary.
Each FDU is represented as a pair of binary adjacency matrices
$(\varphi, \psi)\in\{0,1\}^{3\times3}\times\{0,1\}^{3\times3}$, with rows as targets and
columns as sources, where $\varphi_{ij}=1$ denotes an activating influence $j\to i$,
$\psi_{ij}=1$ denotes an inhibiting influence $j\to i$, and $\varphi_{ij}=\psi_{ij}=0$
denotes no influence.
No directed edge is simultaneously activating and inhibiting
($\varphi_{ij}=\psi_{ij}=1$ is prohibited).

The FDU dictionary's triadic component is grounded in the simplest unambiguous structural
primitives: signed tournaments, three-node patterns with exactly one directed signed edge
per unordered pair, no mutual connections, and no self-regulation.
Each unordered pair admits four possibilities (two directions, two signs), yielding
$4^3=64$ triadic tournaments.
Auto-regulation is captured by six unary self-motifs: three auto-activating (one per node,
with $\varphi_{ii}=1$ and all off-diagonal entries zero) and three auto-inhibiting (one per
node, with $\psi_{ii}=1$ and all off-diagonal entries zero).
The FDU dictionary therefore contains 70 motifs (64 triadic tournaments $+$ 6 unary
self-motifs).

This dictionary is constructive: more general triadic architectures can be assembled by
selecting a subset of FDUs and taking the edge-wise union of their signed adjacencies.
For a set $\mathcal{K}$ of FDUs $\{(\varphi^{(k)}, \psi^{(k)})\}_{k\in \mathcal{K}}$,
the superposition is defined as
\begin{equation}
  \varphi = \bigvee_{k\in \mathcal{K}}\varphi^{(k)}, \qquad
  \psi    = \bigvee_{k\in \mathcal{K}}\psi^{(k)}
\end{equation}
where $\vee$ denotes elementwise logical OR (equivalently, elementwise summation followed
by thresholding at $>0$).
Sign-consistent superpositions require that no directed edge is assigned both activating
and inhibiting roles across the selected FDUs.
Under this restriction, the same union operation applies directly to ternary motif codes:
zeros are filled by nonzeros, and any overlapping nonzero entries must agree.

The FDU dictionary satisfies a completeness property: any sign-consistent, pairwise-connected
three-node triad, including those with mutual interactions, can be realized as the union of
at most two tournament FDUs
(\hyperref[supp:data]{\textbf{Supplementary Material}}).
Specifically, a triad with no mutual pairs is reconstructed by one tournament FDU; otherwise
two tournament FDUs suffice, resolving each mutual pair in opposite directions while agreeing
on all single-direction (forced) edges.
The number of distinct minimal decompositions is exactly $2^{m-1}$ for a triad with
$m \geq 1$ mutual pairs.
Consequently, $m=1$ yields a unique decomposition, $m=2$ yields exactly two, and $m=3$
yields exactly four.
When multiple minimal decompositions exist, a canonical representative is selected and
additional valid superpositions treated as representational redundancy.

Two cases illustrate the completeness property at its extremes.
The pairwise-connected triad \texttt{010100120} contains a mutual positive interaction on
$\{1,2\}$ and is not itself a tournament.
It admits the two-tournament reconstruction \textbf{(Fig.~\ref{fig:2}a)}:
\begin{equation}
  \texttt{000100120} \vee \texttt{010000120}
\end{equation}
The decomposition is unique: a single mutual pair leaves no representational choice.
As a second illustration, the triad \texttt{011201110}, which contains $m=3$ mutual
unordered pairs, admits exactly $2^{m-1}=4$ distinct minimal decompositions
\textbf{(Fig.~\ref{fig:2}b)}:
\begin{equation}
  \begin{aligned}
    &\texttt{000200110} \vee \texttt{011001000} \\
    &\texttt{000201100} \vee \texttt{011000010} \\
    &\texttt{001200010} \vee \texttt{010001100} \\
    &\texttt{001201000} \vee \texttt{010000110}
  \end{aligned}
\end{equation}
In each case, the two tournament FDUs jointly realize both directions for each mutual pair
while preserving the signs and orientations of all forced edges, yielding the same target
triad under sign-consistent union.
All four decompositions realize the same target triad; the multiplicity is representational
redundancy, not inferential ambiguity, and canonical selection handles it in practice
(\hyperref[supp:data]{\textbf{Supplementary Material}}).

Self-regulation extends naturally to the same construction because unary self-motifs operate
exclusively on diagonal entries, which triadic FDUs leave at zero: the two components are
structurally orthogonal, and their union cannot introduce sign conflicts.
The corresponding unary self-motifs are incorporated by the same union operation, leaving
the superposition and sign-consistency rules unchanged.
Auto-inhibition on component 3, for instance, converts \texttt{000102100} to
\texttt{000102102} \textbf{(Fig.~\ref{fig:2}c)}:
\begin{equation}
  \texttt{000102100} \vee \texttt{000000002}
\end{equation}

Together, these results establish that signed triadic interaction structure is
compositionally closed: every admissible three-node configuration is exactly recoverable
from a sparse combination of the simplest unambiguous signed primitives.
The FDU dictionary thus constitutes a complete and exact language for signed three-node
interaction architecture.

\subsection{The Signed Three-Node Interaction Space Partitions into Five
Permutation-Invariant Structural Classes}

A complete enumeration characterizes the signed triad primitive space but does not reveal
whether the 512 signed configurations reduce to a compact set of qualitative structural
types.
Having established this space exhaustively, we partitioned its elements into
permutation-invariant, sign-sensitive structural families.
This taxonomy assigns each triad to a permutation-invariant structural class
within two primary branches, parallel-path ($P$) and single-path ($S$), each further
subdivided by coherence state or feedback polarity.

The classification builds on two matrix representations of the canonical triad encoding
$T\in\{0,1,2\}^{3\times3}$ (rows: targets; columns: sources): a signed adjacency
$A\in\{-1,0,+1\}^{3\times3}$ and an unsigned adjacency $U\in\{0,1\}^{3\times3}$:
\begin{equation}
  A_{ij} = \begin{cases}
    +1, & T_{ij}=1 \\
    -1, & T_{ij}=2 \\
    0,  & T_{ij}=0
  \end{cases}
  \qquad\text{and}\qquad
  U_{ij} = \mathbf{1}[T_{ij}\neq 0].
\end{equation}
Under this convention, $A_{ij}$ is the sign on the directed edge $j\to i$, and $U_{ij}$
indicates whether $j\to i$ exists, irrespective of sign.
The matrix $U^2 = UU$ counts directed length-2 walks from source $j$ to target $i$, so
that $(U^2)_{ij}$ detects whether a direct influence $j\to i$ is accompanied by a two-step
relay $j\to k\to i$.
In a three-node system, $(U^2)_{ij}\in\{0,1\}$ for all $i\neq j$.

Using $(U, U^2)$, parallel-path motifs (class $P$) are those for which at least one ordered
pair ($j\to i$) admits both a direct edge and a two-step relay in parallel:
\begin{equation}
  \exists\, i\neq j \;\text{s.t.}\; U_{ij}=1 \;\text{and}\; (U^2)_{ij}\geq 1.
\end{equation}
In this case, influence from $j$ to $i$ decomposes into a direct arm ($j\to i$) and an
indirect arm through the relay node $k$ (uniquely the third node in a three-node system, $j\to k\to i$).
The feed-forward triad \texttt{000100110} is a canonical example: node 1 regulates node 3
both directly $(1\to3)$ and indirectly via node 2 ($1\to2\to3$), yielding $U_{31}=1$ and
$(U^2)_{31}=1$.

Within class $P$, coherence is determined by whether the direct and relayed contributions
agree in sign on each parallel ordered pair.
For a parallel ordered pair $j\to i$ with relay node $k\neq i,j$, the direct and relay
signs are:
\begin{equation}
  \begin{split}
    \mathrm{sgn}_\mathrm{direct}(j\to i) &= A_{ij},\\
    \mathrm{sgn}_\mathrm{relay}(j\to i)  &= A_{kj}\,A_{ik}.
  \end{split}
\end{equation}
where $\mathrm{sgn}_\mathrm{relay}$ is the sign of the two-step path $j\to k\to i$
(product of edge signs along the path).
A motif is coherent on $j\to i$ if $\mathrm{sgn}_\mathrm{direct}=\mathrm{sgn}_\mathrm{relay}$,
and incoherent otherwise.
The feed-forward triad \texttt{000100110} is coherent, because all nonzero entries of $A$
are $+1$.
In contrast, the related class $P$ motif \texttt{000100120}, in which the $2\dashv3$ arm is
inhibiting rather than activating, is incoherent on the ordered pair $1\to3$: the direct
arm is activating ($A_{31}=+1$), whereas the relayed arm is inhibiting
($A_{21}A_{32}=(+1)(-1)=-1$).

Collectively, class $P$ partitions into three permutation-invariant subclasses according to
the coherence pattern across parallel ordered pairs: $P_c$: coherent on all parallel ordered
pairs; $P_\mathrm{in\text{-}c}$: incoherent on all parallel ordered pairs; and $P_x$:
mixed, containing at least one coherent and at least one incoherent parallel ordered pair.
For example, \texttt{000201220} is mixed: it is coherent on $1\to2$ (direct $A_{21}=-1$;
relay $A_{31}A_{23}=(-1)(+1)=-1$), but incoherent on $1\to3$ (direct $A_{31}=-1$; relay
$A_{21}A_{32}=(-1)(-1)=+1$).

Class $S$ triads are exactly the signed directed 3-cycles: triads in which each unordered
pair carries a single directed edge and these edges together form a cycle.
Formally, class $S$ is defined by the absence of any parallel direct-relay configuration:
\begin{equation}
  \forall\, i\neq j:\; \neg\!\left(U_{ij}=1 \;\text{and}\; (U^2)_{ij}\geq 1\right).
\end{equation}
In these motifs, influence may propagate through chains or cycles, but no ordered pair
$j\to i$ ever has both a direct edge and a two-step route $j\to k\to i$ simultaneously,
a condition that forces this tournament structure as the unique admissible architecture.

Within class $S$, triads are further stratified by feedback polarity, characterized by the
signs of simple directed cycles.
For an oriented 3-cycle $i\to j\to k\to i$, the cycle sign is the product of edge signs
along the cycle:
\begin{equation}
  \mathrm{sgn}(i\to j\to k\to i) = A_{ji}\,A_{kj}\,A_{ik},
\end{equation}
since edge $i\to j$ corresponds to entry $A_{ji}$ under our target-by-source convention.
A class $S$ triad is $S_-$ if any simple directed cycle has negative sign, and $S_+$
otherwise.
A positive-feedback single-path motif is exemplified by \texttt{001100010}, in which the
components form a directed 3-cycle $1\to2\to3\to1$ with all edges activating, yielding only
positive simple cycles.
A negative-feedback single-path motif is given by \texttt{002100010}, which
implements the cycle $1\to2\to3\dashv1$ with net negative sign.

Together, the parallel-path versus single-path split, the coherence state within class $P$,
and the feedback polarity within class $S$ define a permutation-robust taxonomy that assigns
each triad to exactly one of five structural classes \textbf{(Fig.~\ref{fig:2}d,e)}:
\begin{equation}
  \{P_c,\; P_\mathrm{in\text{-}c},\; P_x,\; S_+,\; S_-\}.
\end{equation}

The distribution of all 512 admissible triads reveals a strong
structural skew \textbf{(Fig.~\ref{fig:2}f)}.
Only 16 triads ($3.1\%$) fall into the single-path class $S$, split evenly between
positive-feedback and negative-feedback architectures (8 in $S_+$, 8 in $S_-$).
By contrast, 496 triads ($96.9\%$) are parallel-path motifs (class $P$), containing at
least one ordered pair for which a direct edge and a two-step route coexist in parallel.
Within class $P$, coherent and incoherent motifs occur in equal numbers (124 each in $P_c$
and $P_\mathrm{in\text{-}c}$), while the largest group comprises mixed motifs (248 in
$P_x$) that contain both coherent and incoherent parallel pairs within the same triad.

Two structural features follow.
First, parallel direct-relay structure is combinatorially generic: direct and relay routes
coexist in all but the 16 class $S$ configurations.
Second, the predominance of $P_x$ (248 motifs) indicates that structural heterogeneity,
containing both coherent and incoherent parallel ordered pairs, is common within the class
$P$ space.
Collectively, this invariant-based partition provides a compact, permutation-robust summary
of local three-node architecture that explicitly resolves whether each motif contains
parallel direct-relay channels and, when present, whether those channels align or oppose in
sign (\hyperref[supp:data]{\textbf{Supplementary Material}}).

\subsection{Signed Triad Structure Dictates Minimal Perturbation Panels
for Relay Cancellation}

In directed signed networks, influence between two nodes can propagate simultaneously along
a direct edge and a relay path through an intermediate node, so that a change in a source
node may elicit a composite response at the target that conflates both contributions.
This conflation is not a modeling artifact; it is a structural property of the triad.
Separating these contributions is therefore a prerequisite for structural identifiability.
We asked, for a given signed triad, what minimal perturbation panel achieves this separation
on every parallel pair.
We show that the answer is determined by the triad's local signed structure.

The analysis proceeds around a reference state: responses are locally linear in perturbation
amplitude, and only the sign structure of couplings governs panel design.
We consider three primitive perturbation regimes that can be assembled into perturbation
panels: (i)~single-component perturbations, in which one node is perturbed at a time,
$\mathbf{u}=\pm u_0\,\mathbf{e}_j$; (ii)~opposite-sign co-perturbations, in which two nodes
are perturbed with matched magnitude and opposite polarity,
$\mathbf{u}=\pm u_0(\mathbf{e}_j - \mathbf{e}_k)$; and (iii)~same-sign co-perturbations,
in which two nodes are perturbed with matched magnitude and the same polarity,
$\mathbf{u}=\pm u_0(\mathbf{e}_j + \mathbf{e}_k)$.
Here $\mathbf{u}\in\mathbb{R}^3$ is the perturbation vector with components $u_j$ at node
$j$, $u_0>0$ is the scalar perturbation amplitude, and $\mathbf{e}_\ell\in\mathbb{R}^3$ is
the $\ell$-th standard basis vector ($1$ in coordinate $\ell$, $0$ otherwise).
``Matched magnitude'' means the two nonzero components have equal absolute value
($|u_j|=|u_k|$), so that polarity (same versus opposite sign) is the only additional degree
of freedom introduced by co-perturbation.
Under this matched-amplitude constraint, signed triad structure alone determines the panel assignment.

Fix a parallel ordered pair ($j\to i$) for which influence can propagate both directly and
indirectly within the triad.
In a three-node motif, the relay node $k$ is uniquely determined as the remaining node
distinct from $i$ and $j$.
Using the signed adjacency convention (rows are targets, columns are sources), define the
signed edge indicators along the direct and relayed arms by
\begin{equation}
  \begin{aligned}
    a &\equiv A_{kj} \in\{\pm1\} \quad (\text{sign of } j\to k), \\
    b &\equiv A_{ik} \in\{\pm1\} \quad (\text{sign of } k\to i), \\
    c &\equiv A_{ij} \in\{\pm1\} \quad (\text{sign of } j\to i).
  \end{aligned}
\end{equation}
For such a parallel ordered pair, all three quantities are nonzero.
Under additive sign-only linearization (magnitudes absorbed into a local scaling), the relay
node responds to perturbations at $j$ and $k$ as
\begin{equation}
  \Delta x_k \approx a\,u_j + u_k,
\end{equation}
and the response at $i$ decomposes into a direct term plus a relay term:
\begin{equation}
  \Delta x_i \approx c\,u_j + b\,\Delta x_k \approx c\,u_j + b(a\,u_j + u_k).
\end{equation}
Choosing $u_k$ to enforce $\Delta x_k\approx0$ suppresses the relayed arm, leaving
$\Delta x_i \approx c\,u_j$: the sign of the response at $i$ directly encodes the sign of
the direct edge $j\to i$, converting a composite response into a signed edge readout.
The relay-cancellation condition is therefore
\begin{equation}
  u_k = -a\,u_j.
\end{equation}
Under matched-amplitude co-perturbations ($|u_j|=|u_k|$), this reduces to a polarity rule:
when $a=+1$ (activating $j\to k$), cancellation requires opposite-sign co-perturbation
$u_k=-u_j$, whereas when $a=-1$ (inhibiting $j\dashv k$), cancellation requires same-sign
co-perturbation $u_k=+u_j$.
Notably, the required polarity depends only on the sign of the first leg $j\to k$ and not
on whether the parallel ordered pair is coherent or incoherent.
The design criterion is structurally leaner than the coherence criterion: coherence depends
on the relationship among all three local edge signs ($a$, $b$, $c$), whereas relay
cancellation depends on $a$ alone.
The relay-cancellation polarity is thus fully encoded in a single signed quantity, the first
relay leg $a = A_{kj}$, making panel design a direct read-off from local signed triad
structure \textbf{(Fig.~\ref{fig:3})}.

Applied across the five structural classes, this condition yields three design rules.

\noindent\textbf{Design Rule 1.}
In class $S_+$ and $S_-$ motifs, no ordered pair has a direct edge coexisting with a two-step relay.
Because relay interference is structurally absent, single-component perturbations are
sufficient to probe all signed routes within the motif
(\textbf{Fig.~\ref{fig:3}}: rows~1--2).

\noindent\textbf{Design Rule 2.}
In class $P_c$ and $P_\mathrm{in\text{-}c}$ motifs, the co-perturbation polarity required
for relay cancellation is fully determined by the sign of the first relay leg $A_{kj}$:
opposite-sign co-perturbation ($u_k = -u_j$) when $A_{kj}=+1$; same-sign co-perturbation
($u_k = +u_j$) when $A_{kj}=-1$.
When multiple parallel ordered pairs within the same motif carry opposing first-leg signs,
both co-perturbation polarities must be included
(\textbf{Fig.~\ref{fig:3}}: rows~3--5).

\noindent\textbf{Design Rule 3.}
In class $P_x$ motifs, direct and relayed contributions reinforce on coherent parallel
ordered pairs and oppose on incoherent ones.
Supporting attribution across this mixed coherence structure requires both co-perturbation
polarities: together they yield a controlled relay-off and relay-on contrast, with each
polarity selectively suppressing or amplifying the relayed arm depending on the pair,
supporting robust attribution of direct and relayed contributions across all parallel ordered
pairs within the motif (\textbf{Fig.~\ref{fig:3}}: row~6).

Together, these rules define four minimal panel categories for each signed triad:
single-component, opposite-sign only, same-sign only, or dual-polarity.
Here ``minimal'' refers to the least demanding regime sufficient, under the local
linear-response and matched-amplitude assumptions, to cancel relayed contributions on all
parallel ordered pairs within the motif and to support attribution across all coherence
configurations present.

All 512 admissible signed triads are assigned to exactly one panel category
\textbf{(Table~\ref{tab:panel_assignments}; \hyperref[supp:data]{Supplementary Material})}.
All 16 single-path motifs (8 $S_+$ and 8 $S_-$) fall into the single-component category.
The remaining 496 motifs are parallel-path motifs (class $P$).
Within $P_c$, 52 motifs require only opposite-sign co-perturbations, 51 require only
same-sign co-perturbations, and 21 require dual-polarity panels.
The incoherent subclass $P_\mathrm{in\text{-}c}$ exhibits the mirror pattern (51, 52, and
21, respectively).
The 21 dual-polarity motifs within each of $P_c$ and $P_\mathrm{in\text{-}c}$ arise from
the Rule 2 extension: these motifs contain multiple parallel ordered pairs whose first
relay legs carry opposite signs, generating conflicting polarity requirements despite uniform
coherence across all pairs.
For example, \texttt{000102220} (class $P_c$) has two coherent parallel ordered pairs with
opposing first-leg signs: $a=-1$ on $1\to2$ via $3$, requiring same-sign co-perturbation,
and $a=+1$ on $1\to3$ via $2$, requiring opposite-sign co-perturbation; both polarities
must therefore be included.
Mixed motifs $P_x$ comprise the most demanding regime: all 248 motifs
require dual-polarity panels.
Across all 496 class $P$ motifs, 103 require only opposite-sign co-perturbations, 103
require only same-sign co-perturbations, and 290 ($\approx58\%$) require dual-polarity
panels.
Dual-polarity co-perturbation is thus the most common requirement, applying to 290 of 512
($57\%$) of all admissible signed triads.

The structure-derived panel assignment converts the five-class structural taxonomy into
actionable intervention conditions required by the physics-informed inference model.

\subsection{FDU-Regularized Continuous-Time Dynamics Constitute a Constructive
Physics-Informed Framework for Signed Motif Inference}

Four defining properties embed structural commitment throughout the parameterization: a
constructively complete FDU bank spanning all signed structural primitives across $N$-node
systems; an attention mechanism that concentrates structural attribution onto a compact set
of FDU primitives; a topology-magnitude factorization separating which interactions exist
from how strong they are; and a physics-informed dynamical model that maps structural
parameters to sign-interpretable, operating-point-aware dynamics.

A system of $N$ interacting nodes has state $x(t)\in\mathbb{R}^N$, where $x_i(t)$
denotes the state of node $i$ at time $t$.
Signed directed influence is parameterized through two nonnegative strength matrices
\begin{equation}
  \alpha,\beta\in\mathbb{R}_{\geq0}^{N\times N},
\end{equation}
where $\alpha_{ij}$ encodes the strength of an activating influence $j\to i$ and
$\beta_{ij}$ encodes the strength of an inhibiting influence $j\dashv i$.
Under the target-source convention (rows are targets, columns are sources), the signed
interaction pattern implied by $(\alpha,\beta)$ is directly comparable to the triadic
encodings (activating channel supported where $\alpha_{ij}>0$, inhibiting channel supported where
$\beta_{ij}>0$).

Motif regularization of $(\alpha,\beta)$ is achieved through a finite bank of $N\times N$
binary masks, constructed by lifting the primitive FDUs into the $N$-node setting: each
triadic FDU is embedded onto node triplets (placing its $3\times3$ pattern into the
$N\times N$ mask at the selected rows and columns, zeros elsewhere), and each unary
self-motif onto individual nodes, yielding $\{(M_\uparrow^{(\ell)}, M_\downarrow^{(\ell)})\}_{\ell=1}^{L}$,
where $\uparrow$ indexes activating and $\downarrow$ indexes inhibiting edge templates,
and $L=L(N)$ is finite for any fixed $N$ \textbf{(Fig.~\ref{fig:4}a)}.
Because the FDU dictionary is constructively complete, the bank provides a sufficient
generative vocabulary: any signed triadic architecture in the target network is expressible
as an attention-weighted superposition of bank entries.
The bank size is given by:
\begin{equation}
  L(N) = 64\cdot\binom{N}{3} + 2N,
\end{equation}
where the first term counts placements of the 64 tournament FDUs across all 3-node subsets
and the second accounts for self-regulation at each of the $N$ nodes (activating and
inhibiting).
At $N=3$, $L(3)=70$: exactly the FDU dictionary.
At $N=4$, $L(4)=264$; at $N=20$, $L(20)=73{,}000$, reflecting the polynomial $O(N^3)$
growth of the hypothesis space with network size.
Each mask pair specifies which directed edges are admissible as activating versus inhibiting
under motif $\ell$.
In networks containing singletons, the bank is augmented with a null entry carrying zero masks.

Attention over this bank is induced by the entmax transform
\textbf{\cite{Peters2019entmax}}, a sparsity-promoting normalization $\omega(\cdot)$ that
maps learnable logits $z\in\mathbb{R}^L$ to simplex-constrained FDU attention weights
$e\in\mathbb{R}^L$:
\begin{equation}
  e = \omega(z), \qquad e_\ell\geq0, \qquad \sum_{\ell=1}^{L}e_\ell=1.
\end{equation}
This induces motif-level sparsity that (i)~discourages diffuse superpositions over many
motifs, thereby regularizing the combinatorial hypothesis class, and (ii)~yields a
parsimonious interpretation of the inferred structure as a small set of dominant FDU
primitives whose edge-wise union supports the effective signed interaction pattern
\textbf{(Fig.~\ref{fig:4}b)}.

These weights define motif-derived structural support masks
\begin{equation}
  \sigma_\uparrow   = \sum_\ell e_\ell\,M_\uparrow^{(\ell)}, \qquad
  \sigma_\downarrow = \sum_\ell e_\ell\,M_\downarrow^{(\ell)},
\end{equation}
where $(\sigma_\uparrow)_{ij}$ and $(\sigma_\downarrow)_{ij}$ quantify how strongly the
current FDU mixture supports activating or inhibiting influence on edge $j\to i$.
Entry-wise gating by the support masks yields the effective signed interaction strengths
used by the dynamics:
\begin{equation}
  \alpha^\mathrm{eff} = \alpha \odot \sigma_\uparrow, \qquad
  \beta^\mathrm{eff}  = \beta  \odot \sigma_\downarrow,
\end{equation}
with $\odot$ denoting Hadamard multiplication.
In this factorization, FDUs act as structural priors that constrain signed topology through
$(\sigma_\uparrow, \sigma_\downarrow)$, while $(\alpha,\beta)$ encode magnitudes for the
admissible interactions.
This factorization separates two logically distinct inference problems: which interactions
exist (topology, determined by FDU attention) and how strong they are (magnitude, encoded
in $\alpha,\beta$), allowing each to be learned through its natural parameterization
\textbf{(Fig.~\ref{fig:4}c)}.
Because $(\sigma_\uparrow, \sigma_\downarrow)$ arise from superposition over a finite FDU
dictionary, the parameterization can express composite architectures not represented by any
single motif, while remaining anchored to interpretable building blocks.

This parameterization acquires continuous-time semantics through an ODE governing the state
$x_i(t)$ of each node, $i=1,\ldots,N$:
\begin{equation}
  \frac{dx_i(t)}{dt} = F_i(x(t)) - \gamma_i\,x_i(t) + u_i(t),
\end{equation}
where $F_i$ is the net signed interaction drive, $\gamma_i>0$ is a first-order decay rate,
and $u_i(t)$ is an exogenous input representing a controlled perturbation applied to
node $i$.
The controlled inputs $u_i(t)$ are instantiated as the relay-cancellation panel prescribed
by design rules 1--3, guaranteeing that training data spans the relay-off and relay-on
conditions necessary for signed edge attribution under the structural class of the target
motif.
The physics-informed constraint that $x(t)$ satisfies the ODE at every observed time point
reduces structural degeneracy: where multiple interaction structures may fit isolated
observations, only a subset generate ODE-consistent trajectories across the full
perturbation panel.

$F_i$ is an additive superposition of activating and inhibiting channels from all sources,
capturing saturating nonlinear effects while preserving sign interpretability:
\begin{equation}
  F_i(x) = \sum_{j=1}^{N}\!\left(
    \alpha_{ij}^\mathrm{eff}\,H_\uparrow(x_j;\kappa_j,K_j)
    + \beta_{ij}^\mathrm{eff}\,H_\downarrow(x_j;\kappa_j,K_j)
  \right),
\end{equation}
where $\alpha_{ij}^\mathrm{eff}\geq0$ and $\beta_{ij}^\mathrm{eff}\geq0$ are the
FDU-gated effective strengths for activating and inhibiting influence on the directed edge $j\to i$.
Here $H_\uparrow$ is monotone increasing and saturating, whereas $H_\downarrow$ is monotone
decreasing and saturating.
For systems with nonnegative states, a standard choice \textbf{\cite{Hill1910}} is Hill-type forms:
\begin{equation}
  \begin{split}
    H_\uparrow(x;\kappa,K) &= \frac{x^\kappa}{K+x^\kappa}, \\
    H_\downarrow(x;\kappa,K) &= \frac{K}{K+x^\kappa},
  \end{split}
\end{equation}
with $\kappa_j\geq1$ controlling cooperativity and $K_j>0$ setting the half-saturation
scale for source node $j$.
The saturating Hill forms directly address operating-point dependence: when a source node's
state lies far from the half-saturation point $K$, the Hill response is near-maximal or
near-zero, and even a structurally present interaction contributes negligible drive to $F_i$.
Trajectory-spanning training across perturbation conditions is therefore the operative
mechanism: transient phases in which the source traverses the sensitive regime near $K$
carry the discriminative signal for structural recovery.
Under this parameterization, $\partial H_\uparrow/\partial x>0$ and
$\partial H_\downarrow/\partial x<0$ for $x>0$, so edges with $\alpha_{ij}^\mathrm{eff}>0$
and $\beta_{ij}^\mathrm{eff}=0$ induce $\partial F_i/\partial x_j>0$ (small-signal
activation), whereas edges with $\beta_{ij}^\mathrm{eff}>0$ and $\alpha_{ij}^\mathrm{eff}=0$
induce $\partial F_i/\partial x_j<0$ (small-signal inhibition)
\textbf{(Fig.~\ref{fig:4}d)}.
The parameterization is trained to satisfy channel exclusivity: each directed edge operates
in at most one channel.
Under channel exclusivity, the sign of $\partial F_i/\partial x_j$ directly encodes
the signed interaction $j\to i$, grounding the learned dynamics in the motif representation.

This framework resolves all three coupled inferential challenges: a tractable
interaction-structure hypothesis space, a structure-derived perturbation design, and a
physics-informed, state-dependent mechanistic dynamical model.

\subsection{FDU-Regularized Physics-Informed Inference Recovers Signed Structure,
Motif Identity, and Dynamics Without Per-Edge Structural Labels}

The FDU-regularized parameterization provides a tractable and constructively complete
hypothesis class for signed
motif inference; its empirical adequacy is assessed against representative motifs spanning
all five structural classes.
We asked whether the inferred $(\alpha,\beta)$ channels recover edge presence,
source-to-target directionality, and activating or inhibiting polarity, whether the learned
dynamics faithfully regenerate perturbation-resolved trajectories, and whether the FDU attention mechanism localises onto the generating motif's support set.

The six representative motifs \textbf{(Fig.~\ref{fig:3})} span all five structural classes
and all three perturbation panel types prescribed by design rules 1--3: \texttt{002100010}
(class $S_-$),
\texttt{001100010} (class $S_+$), \texttt{000100110} (class $P_c$), \texttt{000200210}
(class $P_c$), \texttt{000100120} (class $P_\mathrm{in\text{-}c}$), and \texttt{000201220}
(class $P_x$).
The two class $P_c$ representatives cover both the opposite-sign and same-sign
co-perturbation sub-cases, spanning the full range of panel types within that class.
For each motif, ground-truth trajectories were generated by simulation from the
FDU-regularized continuous-time system with known structure, under the unperturbed reference
condition and the class-dependent perturbation panel: class $S$ motifs (\texttt{002100010},
\texttt{001100010}) received single-component panels; the two class $P_c$ motifs received
opposite-sign (\texttt{000100110}) and same-sign (\texttt{000200210}) co-perturbation panels
respectively; the class $P_\mathrm{in\text{-}c}$ motif (\texttt{000100120}) received an
opposite-sign co-perturbation panel; and the class $P_x$ motif (\texttt{000201220}) received
a dual-polarity panel.
Training and validation used sparse observations distributed across distinct dynamical phases
(initialization, early response, mid-to-late transient, and steady state; 3:1
training-to-validation split).
Each motif was trained jointly across all perturbation conditions in its class-specific
panel under a fixed protocol with identical hyperparameters across experiments (see Methods).

Structural recovery is assessed from the inferred activating and inhibiting channels
(\textbf{Fig.~\ref{fig:5}a--f}: top row of each panel).
Across the six motifs, dominant interactions concentrate on the ground-truth ordered edges:
activating couplings in $\alpha$ and inhibiting couplings in $\beta$, with non-edges
remaining suppressed in both channels.
This channel separation establishes recovery of edge presence, source-to-target
directionality, and activating-or-inhibiting polarity directly from $(\alpha,\beta)$,
without additional sign assignment beyond the two-channel parameterization.
This signed partition emerges from training rather than explicit supervision: the assignment
of interactions to activating or inhibiting channels is not prescribed per edge, but arises
through the physics-informed dynamics and soft channel exclusivity constraint across all six
motifs.

Trajectory regeneration is evaluated across all six motifs and their class-specific
perturbation panels \textbf{(\hyperref[supp:fig1]{Supplementary Fig.~1})}.
Predicted trajectories closely track ground-truth simulations over the full simulated time
horizon, capturing transient responses and late-time stabilization under the reference
condition, single-component perturbations, and class-specific co-perturbations.
Reconstruction error is quantified on the held-out test set spanning the full temporal
horizon, on time points not used in training or validation, per condition and per state as
the range-normalised root-mean-square error
(\textbf{Fig.~\ref{fig:5}a--f}: row~2 left): the median lies in the range $0.12$ to $0.21$
across motifs, indicating that predicted dynamics remain faithful to ground-truth variability
across all conditions and states.
Training loss curves decrease and stabilize across all experiments
(\textbf{\hyperref[supp:fig1]{Supplementary Fig.~1}}: bottom row), supporting convergence of the joint
trajectory-structure inference.
These results confirm that the inferred continuous-time model generalises across the full
temporal horizon from sparse multi-phase supervision, rather than fitting isolated dynamical
phases.
The simultaneous recovery of accurate trajectories and correct signed structure indicates
that the structural constraints imposed by FDU regularization do not compromise predictive
fidelity.

Motif identification is assessed from the FDU attention distributions over the bank
(\textbf{Fig.~\ref{fig:5}a--f}: row~2 right).
For the five motifs with a unique single-FDU representation, attention mass concentrates on
the generating triad encoding, with markedly smaller weights on alternatives.
For the class $P_x$ motif \texttt{000201220}, attention follows the anticipated compositional
pattern: dominant weight distributes across the two FDUs \texttt{000201200} and
\texttt{000200220}, consistent with the unique minimal two-FDU decomposition for a motif
with a single mutual pair (\hyperref[supp:data]{\textbf{Supplementary Material}}).
Together, these patterns confirm that the attention mechanism localises onto the FDU support
set consistent with the generating architecture, stable across class-dependent perturbation
panels.
The FDU attention profile thus provides a structural basis for the recovered interaction
magnitudes: beyond localising onto the correct motif, it identifies the structural primitive
underlying the inferred interaction pattern, grounding structural recovery in the motif
representation rather than in interaction magnitudes alone.

Element-wise convergence of $\alpha_{ij}$ and $\beta_{ij}$ across training epochs provides
a mechanistic audit of identifiability (\textbf{Fig.~\ref{fig:5}a--f}: row~3).
Entries corresponding to true directed interactions increase and stabilize in the appropriate
channel, whereas coefficients for non-edges remain near baseline or decay toward zero.
In motifs containing inhibiting interactions, growth concentrates in $\beta$ for the
relevant ordered pairs while $\alpha$ remains suppressed, yielding explicit polarity
separation at the parameter level.
Sparse signed structure thus emerges through selective parameter stabilization rather than
diffuse weight sharing across candidate edges.

Beyond the six triad-only experiments, compositional structure recovery is tested on motif
\texttt{000102102} \textbf{(\hyperref[supp:fig2]{Supplementary Fig.~2})}, which comprises the triadic architecture
\texttt{000102100} (class $P_\mathrm{in\text{-}c}$) and a unary auto-inhibiting self-edge
(\texttt{000000002}).
Trajectories are generated under the reference condition and the opposite-sign
co-perturbation panel of its $P_\mathrm{in\text{-}c}$ base triad, under the same fixed
protocol as the triad-only experiments.
The inferred activating and inhibiting channels recover the self-edge in the appropriate
channel while maintaining sparsity on unrelated ordered pairs.
Attention localises to two dominant entries: the triadic component \texttt{000102100} and
the unary self-motif \texttt{000000002} (both present in the bank), autonomously discovering
the two-component FDU decomposition from trajectory data without per-edge structural labels.
Element-wise parameter trajectories show selective stabilization of the self-coupling
alongside the triadic interactions.
Trajectory regeneration remains accurate across all perturbation conditions with a median
range-normalised root-mean-square error of $0.19$ and no evidence of condition-specific
degradation, and training loss decreases and stabilizes consistently.
The compositional FDU representation thus extends seamlessly to composite triadic-unary
architectures: structural complexity is absorbed without sacrificing predictive fidelity or
channel interpretability.

Collectively, across all five structural classes and a compositional architecture extending
FDU coverage to a unary auto-inhibiting self-edge, the FDU-regularized physics-informed neural ODE
jointly recovers signed interaction structure and perturbation-resolved dynamics from sparse
time-course data.
Three forms of structural recovery emerge without per-edge structural labels: assignment
of activating and inhibiting interactions to their respective channels, localisation of FDU
attention onto the generating motif support set, and autonomous discovery of compositional
FDU decompositions.
Structural constraints and predictive fidelity are not in tension: the jointly inferred
solutions remain accurate across the full perturbation panel and consistent with the
ground-truth dynamical system.
These results establish that structural inductive biases constitute the operative inferential
mechanism for mechanistically interpretable signed motif inference from perturbation time
courses.

\subsection{FDU-Regularized Inference Recovers Signed Interaction Structure
of a Network-Embedded Triad Under Open-System Confounding}

Structural inference within a self-contained three-node system rests on a closure
assumption: the perturbation panel covers all nodes, and the local dynamics are fully
determined by intra-block interactions alone.
The relay-cancellation argument exploits this closure directly: for any ordered pair
$(i\to j)$ in a three-node motif, the only possible two-step relay is mediated by the
third node, making relay cancellation a unique, structurally determined operation.
In networks with $N>3$, this closure fails: even when inference is restricted to a local
motif, its observed dynamics can be strongly shaped by influences from nodes outside the
local block that provide alternative relay routes and exogenous drives with no representation
in the closed local model.
When this external contribution is unmodeled, it is generically projected onto intra-block
edges, producing spurious couplings indistinguishable from true local regulation under
standard fitting.
The open-system partition of the network state makes this confounding structurally explicit
and yields a decomposition of the local dynamics that separates the FDU-representable
structure from the environmental contribution.

The full network state separates into a local block $x_B(t)\in\mathbb{R}^{|B|}$, comprising
the $|B|$ nodes under inference ($B\subset\{1,\ldots,N\}$, with $|B|=3$ for an embedded triad),
and an environment block $x_E(t)\in\mathbb{R}^{N-|B|}$, consisting of all remaining nodes
(environment $E$, the set of all nodes outside $B$):
\begin{equation}
  x(t) = \bigl(x_B(t),\, x_E(t)\bigr).
\end{equation}
The key consequence of this partition is that the true dynamics of each local node
$i\in B$ depend on both blocks:
\begin{equation}
  \frac{dx_i(t)}{dt} = F_i(x_B(t),x_E(t)) - \gamma_i\,x_i(t) + u_i(t),
  \qquad i\in B,
\end{equation}
where $F_i(x_B,x_E)$ is the net signed interaction drive on node $i$ from the full network,
$\gamma_i>0$ is a first-order decay rate, and $u_i(t)$ is a controlled perturbation input.
The $x_E(t)$ dependence in $F_i$ is the structural source of open-system confounding:
environment nodes contribute to the drive on local nodes, and this contribution is not
representable by a closed model of the local block alone.

The $x_E(t)$ dependence across all local nodes $i\in B$ is captured collectively in the
vector-form dynamics of the local block:
\begin{equation}
  \frac{dx_B(t)}{dt} = F_B(x_B(t),x_E(t)) - \gamma_B\,x_B(t) + u_B(t),
\end{equation}
where $F_B(x_B,x_E)\in\mathbb{R}^{|B|}$ is the vector of net signed interaction drives with
$i$-th entry $F_i(x_B,x_E)$ for each $i\in B$, $\gamma_B\in\mathbb{R}^{|B|}$ is the vector of
first-order decay rates with $i$-th entry $\gamma_i$, and $u_B(t)\in\mathbb{R}^{|B|}$ is the
vector of controlled perturbation inputs with $i$-th entry $u_i(t)$.
Even with the full system state available, $F_B$ contains contributions from the environment
$E$ not representable by the FDU structural prior.
$F_B$ therefore admits a decomposition:
\begin{equation}
  F_B(x_B,x_E) = \widetilde{F}_B(x) + w(t),
\end{equation}
where $\widetilde{F}_B(x)\in\mathbb{R}^{|B|}$ captures all signed directed interactions
representable by the FDU prior across the full system state $x$, and $w(t)\in\mathbb{R}^{|B|}$
(the latent forcing term) is the exact remainder between the true interaction field and the
FDU-regularized approximation.
The local block dynamics then become:
\begin{equation}
  \frac{dx_B(t)}{dt} = \widetilde{F}_B(x(t)) - \gamma_B\,x_B(t) + u_B(t) + w(t).
\end{equation}
The term $w(t)$ is therefore not an ad hoc modification; it is the unavoidable remainder
between the true local interaction field and any structured approximation operating in an
open dynamical system \textbf{\cite{Alvarez2009latentforce,Rackauckas2020ude}}.
Because this remainder is defined over the full interaction field without reference to the
structural class of the local block, the open-system decomposition is class-agnostic: the
triad's class determines the perturbation panel required for relay cancellation (design
rules 1--3), but leaves the latent forcing requirement unchanged.

Since $w(t)$ is defined as the residual between the true interaction field and the FDU
approximation, it is not directly observable; estimating it from the perturbation panel
jointly with $\widetilde{F}_B$ is therefore necessary.
While the decomposition holds independently of the triad's structural class, the realised
$w(t)$ is specific to each perturbation condition.
Two structural requirements constrain its form.
First, $w(t)$ is inherently time-varying: since $x_E(t)$ changes throughout a perturbation
experiment, the environmental contribution to local dynamics cannot be captured by a static
offset.
Second, $w(t)$ must be condition-conditioned: each perturbation condition drives a distinct
$x_E(t)$ trajectory, so the environmental drive differs systematically across the
perturbation panel.
A finite Fourier time basis $\Phi(t)$ \textbf{\cite{Brunton2016}} with condition-dependent
learned coefficients provides one estimable parameterization that satisfies both
requirements:
\begin{equation}
  \Phi(t) = \bigl[1,\;
    \sin(2\pi j\, t/\tau),\;
    \cos(2\pi j\, t/\tau)
  \bigr]_{j=1}^{J} \in \mathbb{R}^{1+2J},
\end{equation}
where $\tau>0$ normalizes the time scale and $J$ sets the frequency resolution.
The basis $\Phi(t)$ is smooth and globally defined, bounding $w(t)$ to a
finite-dimensional function space; condition dependence is encoded in learned coefficients
over this basis, allowing $w(t)$ to track the environmental trajectory specific to each
perturbation condition.
This parameterization maintains separability from structural inference: intra-block
interactions encoded in $\widetilde{F}_B$ are time-persistent, structurally organized
features that hold across perturbation conditions; $w(t)$ is smooth, condition-specific,
and carries no directed graph structure.
The two components therefore impose distinct functional constraints on the observed local
dynamics, making joint estimation of signed interaction structure and environmental forcing
tractable from the perturbation panel alone.
Four penalties constrain $w(t)$ during training to maintain this separation:
$\mathcal{L}_\mathrm{energy}$ keeping the forcing minimal,
$\mathcal{L}_\mathrm{smooth}$ enforcing temporal regularity,
$\mathcal{L}_\mathrm{row}$ localising environmental drive to a sparse node subset, and
$\mathcal{L}_\mathrm{align}$ preventing $w(t)$ from absorbing perturbation-driven
structural signal (see Methods).

We asked whether the framework recovers the signed interaction structure of a target triad
embedded within a larger open system.
One architecture commonly observed in empirical directed networks combines global sparsity
with dense local interaction motifs, where individual nodes participate in multiple
overlapping interaction subgraphs spanning distinct structural classes.
The proof-of-concept network encodes this architecture \textbf{(Fig.~\ref{fig:6}a)}: a
20-node directed signed network with 10 overlapping signed triad subnetworks,
comprising 43 directed signed edges across four structural classes
($P_x$: 4; $P_\mathrm{in\text{-}c}$: 3; $S_-$: 2; $P_c$: 1).
The network is globally sparse: no node participates in more than two triad subnetworks and
no node carries more than three signed edges per direction, precluding interaction hubs.
Node 12 (singleton) participates in no triad subnetwork; the FDU bank is augmented with a
null entry to cover such cases, bringing the total to 73,001 candidate placements
($L(20) + 1$).
The inference target is the triad of nodes 1, 16, and 18, a class $P_\mathrm{in\text{-}c}$
triad, with seven directed edges sourced from the three target nodes, five spanning
intra-triad interactions and two outgoing to the broader network.
The triad constitutes an upstream interaction core whose internal dynamics are self-contained
for two of its three nodes: nodes 1 and 18 receive no inputs from outside the triad.
Node 16 is the triad's sole external interface, receiving one activating input from node 4
and propagating inhibiting signals back to nodes 4 and 5.

The triad class and its external connectivity jointly determine the perturbation panel,
which addresses two structural requirements.
Relay cancellation within the $P_\mathrm{in\text{-}c}$ triad requires opposite-sign
co-perturbation on the parallel ordered pairs (design rule 2), generating six conditions
across all three target node pairs in both directions.
Single-component perturbations in both directions on each target node isolate each node's
direct downstream effects, making the outgoing connections to the broader network observable
in the perturbation response; together with a reference condition, the panel comprises 13
conditions, where comprehensive coverage is motivated by structural ignorance prior to
inference.
The perturbation effect profile across all 20 nodes characterizes node 16's role as external
interface: nodes 4 and 5, the direct recipients of node 16's inhibiting output, show the
highest median effect sizes among non-target nodes; nodes 7 and 15, two steps removed
through node 5's outgoing inhibiting connections, show intermediate effects; the remaining
13 non-target nodes exhibit near-zero effects (median $<0.01$), reflecting the triad's
limited external connectivity \textbf{(Fig.~\ref{fig:6}b)}.

Structural recovery at the 20-node scale is assessed from the inferred $(\alpha, \beta)$
channels \textbf{(Fig.~\ref{fig:6}c)}.
All seven directed edges from target nodes 1, 16, and 18 are recovered: the two edges from
node 1 concentrate in the activating channel, the three edges from node 16 concentrate
exclusively in the inhibiting channel, and node 18 contributes one activating and one
inhibiting edge.
Convergence trajectories for the target source columns confirm the channel separation
\textbf{(Fig.~\ref{fig:6}d)}: inhibiting-channel entries separate cleanly between recovered
edges and near-zero non-edge entries; the activating channel shows the same separation,
with non-edge entries at distinctly lower magnitudes than the recovered edges.
Polarity is correctly assigned across all seven edges, and trajectory fidelity remains low
across all perturbation conditions and target nodes, with a median range-normalised root-mean-square error of $0.12$
\textbf{(Fig.~\ref{fig:6}e)}, demonstrating that open-system confounding, when explicitly
modeled via latent forcing, does not degrade inferential precision.

The FDU attention profile \textbf{(Fig.~\ref{fig:6}f)} provides the structural basis for
the recovered interaction magnitudes: each non-zero entry in the $(\alpha, \beta)$ channels
is grounded in specific structural primitives whose attention weights quantify their
contribution, linking recovered edges to a compact set of contributing structural
primitives.
Of the 73,001 candidate placements, 72,995 are suppressed to zero or near-zero weight
(individual weights $< 0.05$): this concentration is not merely a sparsity statistic but
evidence that a hypothesis space of 73,001 structural candidates was correctly navigated,
with a specific structural claim emerging from inference.
The weights within this support are distributed: the class $P_\mathrm{in\text{-}c}$ target
triad has a minimal two-FDU decomposition, precluding concentration on a single entry, and
the surrounding network structure broadens the effective support further.
Both decompositions, (\texttt{000100120}, \texttt{001102000}) and (\texttt{000102100},
\texttt{001100020}), are represented within the top ten FDUs, with all four corresponding
entries appearing at ranks 3, 4, 6, and 8, confirming that attention includes the
structural primitives from which the target triad is built
(\hyperref[supp:data]{\textbf{Supplementary Material}}).
The three highest-weight FDUs outside the target triad's decomposition (ranks 1, 2, and 5:
\texttt{020002100} and \texttt{010000220}, the latter reflecting two distinct node-set
embeddings of the same FDU type) all involve nodes 4 and 5, the most dynamically responsive
non-target nodes in the perturbation effect profile \textbf{(Fig.~\ref{fig:6}b)} and node
16's direct downstream targets \textbf{(Fig.~\ref{fig:6}c)}.
Each carries confirmed edges from the full ground-truth network, establishing that the
attention beyond the target triad's decomposition reflects the network's real interaction
neighborhood rather than residual noise.
The remaining support FDUs each anchor on a single confirmed ground-truth inhibiting edge:
one on the inhibiting edge from node 5 to node 7, and two on the inhibiting edge from node
16 to node 4; each carries two additional edges absent from the ground truth, topologically
forced by the 3-node primitive structure and remaining sub-threshold.
The attention profile therefore encodes two nested levels of structurally grounded
inference: the exact FDU composition of the target triad at the core, and the network's
local interaction structure in the surrounding support.
This structural attribution derives from the FDU hypothesis space itself: because every
recovered interaction is expressed as a weighted combination of structural primitives,
grounding is produced intrinsically by inference rather than by secondary analysis of
interaction magnitudes.

The generality of structural recovery is assessed by extending inference independently
to all ten embedded triads, each with its structure-derived perturbation panel, spanning
all four structural classes and all four panel types present in the network.
The area under the receiver operating characteristic curve (AUROC), pooled across all ten
triad scoring matrices covering both signed channels for the three source nodes against all
20 targets, is $0.998$; 70 of 75 ground-truth-positive channel entries are detected, with
5 false positives (F1 $= 0.93$;
\hyperref[supp:table1]{\textbf{Supplementary Table~1}}).
The five false negatives are all intra-triad edges; four inhibiting interactions and one
activating edge fall just below the detection threshold, reflecting weak signals at the
boundary of detectability rather than structural misidentification.
Outgoing sensitivity is perfect across all ten triads and all four panel types prescribed by
the design rules: every directed edge from a target node to a non-target node is recovered
with correct polarity, confirming that the latent forcing strategy successfully decouples
environmental confounding from outgoing structural parameters.
$\mathcal{L}_\mathrm{align}$ follows the same peak-then-decline trajectory across all
runs, regardless of structural class or panel type
(\hyperref[supp:fig3]{\textbf{Supplementary Fig.~3}}), confirming empirically that the
open-system decomposition and the forcing penalty system are class-agnostic; the peak
magnitude scales with panel complexity, reflecting the larger perturbation-correlated
signal available in richer panels, but all runs converge to a stable low-loss regime
before structural inference is complete.
The remaining three penalties ($\mathcal{L}_\mathrm{energy}$, $\mathcal{L}_\mathrm{row}$,
$\mathcal{L}_\mathrm{smooth}$) exhibit uniform behaviour across all structural classes,
confirming that the penalty design is robust and that no class or panel configuration
places unusual demands on the forcing parameterization.
Disabling the latent forcing module under matched conditions collapses structural
recovery to near chance (AUROC $= 0.60$; F1 $= 0.27$), confirming that the module
is necessary for signed structural inference under open-system confounding.

Together, these results demonstrate that open-system confounding is not a fundamental
barrier to signed structural inference: it is a structurally separable sub-problem whose
resolution is achievable through parameterization that exploits the functional contrast
between the FDU interaction field and the environmental contribution.
\section{Discussion}

Signed interaction inference in networked dynamical systems has a well-known failure mode:
the quantity the data identify and the quantity the model parameterizes live at different
structural levels.
Edge-level sign assignments are under-constrained by trajectory data, but the sign pattern
of a triad is not, and it is the triad, not the edge, that carries the mechanistic
content of the interaction.
Grounding the parameterization at this level resolves the mismatch: the structural
constraints that determine what the data can identify and the inferential constraints that
determine what the model can recover become the same constraints.

The FDU framework rests on a specific reductionist commitment: the atomic level of
representation is determined by the structure of the inference problem itself.
The triad is that level: the minimal closed structure in which direct and relay pathways
coexist between the same node pair.
The triad's structural class determines the perturbation conditions required for signed
edge identifiability, making it the natural primitive for networked dynamical systems in
which multi-path confounding is structurally present.
The FDU dictionary is constructively complete at this level, ensuring that nothing in the
signed interaction space falls outside the representation.

This completeness is the first of four properties that jointly characterize the FDU
dictionary as a well-defined hypothesis space for signed structural inference.
It is complete in the representational sense: every admissible signed triad is exactly
representable as a sign-consistent superposition of at most two tournament FDU primitives.
It is interpretable: each primitive belongs to a defined structural class with explicit
pathway and perturbation-design consequences.
It is self-specifying: the structural class of the target triad directly prescribes the
perturbation conditions sufficient for signed edge identifiability.
It is tractable: the $N$-node bank grows as $O(N^3)$.
Together, these properties define the FDU dictionary as a well-formed hypothesis space.

The interpretability of the FDU hypothesis space follows from a specific shift in
representational granularity.
Where a conventional directed graph specifies edges independently as pairwise adjacency
entries, FDU superposition couples edge choices within each three-node subgraph, making
pathway composition and coherence intrinsic to the representation rather than inferences
drawn from independent entries; this is the shift from surface to volumetric triangulation.
The constructive property of the FDU dictionary makes this concrete: complex interaction
architectures are not special cases requiring bespoke representation, but natural
compositions of the same primitive set.

Every non-zero entry in the learned interaction matrices is structurally mediated through
FDU support: each recovered activating or inhibiting edge is explained by specific FDU
primitives whose attention weights quantify their contribution to the observed dynamics.
This attribution is intrinsic to the inference: it is not derived by comparing recovered
parameter values to an external structural vocabulary after the fact, but by the FDU
attention mechanism, which constrains every learned interaction to remain expressible as a
superposition of interpretable structural units.
In networked systems with multiple overlapping subgraphs, this makes the result structurally
interpretable beyond edge recovery: the FDU attention profile identifies the structural
primitives consistent with each recovered edge and their relative contributions, grounding
inference in the motif representation rather than in interaction magnitudes alone.
The separation of relay-mediated from direct contributions, and the attribution of edge sign at 
the structural level, are properties that follow from the choice of hypothesis space rather than 
from the learning procedure (cf.~\textbf{\cite{Lin2025PerturbODE}}).

Beyond structural attribution, the structure-derived perturbation panel resolves a specific
class of inference errors.
When perturbation design is decoupled from the triad-level structural class of the target
interaction, relay-mediated and direct contributions remain conflated, parallel-path triads
cannot be resolved into their coherent or incoherent configurations, and inferred networks
accumulate spurious direct edges: relay-mediated connections misattributed to direct
influence.
The design rules demonstrate that this decoupling is not inevitable.
The signed structural class of a target triad prescribes the minimal perturbation conditions
required for relay cancellation and direct-arm isolation, making perturbation design a
derived consequence of the structural hypothesis rather than an independent choice.
By construction, the prescribed panel probes the identifiable directions in parameter space
along which structurally competing hypotheses diverge.
This establishes the forward direction: class-to-panel, not panel-to-class.
When the true local structure is unknown, a conservative panel combining single-component
perturbations with dual-polarity co-perturbations provides coverage across all admissible signed triads, preserving the framework's
applicability without requiring prior class knowledge.

The present validation is conducted on synthetic data, a deliberate design choice that
isolates modeling contributions from the challenges inherent in real-world settings;
the framework's robustness to measurement noise, irregular temporal acquisition, model
misspecification, and preprocessing variation remains to be established.
The model also holds the Hill kinetic parameters (cooperativity, half-saturation, and decay
rate) at ground-truth values, isolating structural recovery from the harder problem of joint
parameter estimation; in practice these are unknown and must be inferred jointly with
interaction structure, substantially expanding the effective parameter space.
Another consideration is computational complexity: the $O(N^3)$ bank grows polynomially
with $N$, and computational costs for evaluating attention over the full bank become
non-trivial at larger scales; entmax sparsification mitigates this by concentrating structural attribution
onto a compact active support, reducing the effective inferential complexity regardless of
bank size.
These considerations reflect deliberate scope boundaries of the present proof-of-concept
rather than fundamental barriers, and together constitute a well-defined agenda for
extending the framework toward real-world settings.

Directed signed interactions, targetable perturbations, and time-resolved observations
are the entry conditions for the FDU framework.
They are most directly satisfied in molecular biological systems.
Transcriptional regulatory networks \textbf{\cite{de2002modeling,lee2002transcriptional}}
meet all three: genetic knockouts and overexpression provide targeted node perturbations,
and transcriptomic time courses yield the required observations.
Independently, the coherent and incoherent feed-forward loops that appear as identifiable
subclasses in the structural taxonomy are the same motifs previously identified as
enriched above random expectation in biological regulatory networks
\textbf{\cite{Milo2002}}: a correspondence between what the framework deems inferentially
distinguishable and what biology has selected for.
The entry conditions may also be approximated outside experimental settings, when
structurally localized natural events (social disruptions with a known locus
\textbf{\cite{Leskovec2010signed}}, exogenous economic shocks to identifiable sectors
\textbf{\cite{InoueTodo2019,Acemoglu2012network}}) play the role of targeted
perturbations, and time-resolved observations are available.
Testing these settings is beyond the scope of this work, but the entry conditions and
design rules provide a principled basis for domain-specific adaptation.

The difficulty of signed structural inference has long been read as a data problem.
FDU reframes it as a representational one.
Once the hypothesis space, the perturbation design, and the dynamical model are indexed
by the same primitive, signed structure is not something to be extracted from the data
but something the data and the model agree on by construction.

\section{Methods}

Signed structural inference from perturbation time-series places three distinct computational
demands on the framework: predicting trajectories across the full perturbation panel,
recovering the sparse FDU interaction structure underlying them, and, in the open-system
setting, absorbing contributions that fall outside the FDU representation.
Three modules address these demands: a reference-conditioned trajectory predictor, a
motif-regularized structural parameterizer, and a latent forcing absorber, trained jointly
under a composite objective and a progressive optimization schedule that preserves the
perturbation-response signal on which structural inference depends.
A complete list of mathematical symbols is provided in the
\hyperref[supp:notation]{Supplementary Note}.

\subsection{Model Architecture}

The trajectory predictor maps an input triplet $(t,\, x_\mathrm{ref}(t),\, u)$ to the
predicted state $\hat{x}(t) \in \mathbb{R}^N$, where $t$ is the query time,
$x_\mathrm{ref}(t) \in \mathbb{R}^N$ is the unperturbed reference state at that same time,
and $u \in \mathbb{R}^N$ is the perturbation vector.
$x_\mathrm{ref}(t)$ is obtained by direct evaluation of the unperturbed ODE solution at
each query time.
Conditioning on $x_\mathrm{ref}(t)$ rather than a fixed initial condition at $t = 0$
removes the need to learn baseline dynamics from scratch: perturbation response is a
differential quantity, and providing the reference state as temporal context directs
representational capacity toward the perturbation-driven deviation that carries structural
information.
Because $u$ is an explicit input, a single predictor handles all perturbation conditions
simultaneously without per-condition retraining.

The network is a multilayer perceptron with $\tanh$ activations throughout: their
everywhere-differentiable smoothness is required for exact computation of $d\hat{x}/dt$
through the network.
The exact time derivative $d\hat{x}/dt$ is obtained at any queried time point without
finite-difference approximation and on arbitrarily irregular time grids; it is computed via
forward-mode automatic differentiation, specifically a Jacobian-vector product with unit
direction in the time axis, in a single forward pass through the network.
Layer normalization \textbf{\cite{Ba2016}} and dropout \textbf{\cite{Srivastava2014}}
stabilize optimization under the composite physics-informed objective; weights are initialized from a Xavier uniform distribution, which preserves
signal variance across layers for $\tanh$ nonlinearities \textbf{\cite{Glorot2010}}.

The structural parameterizer separates the inference of interaction topology from the
inference of interaction magnitude.
Topology, the set of directed interactions that are structurally supported, is determined by
entmax-regularized \textbf{\cite{Peters2019entmax}} attention over the FDU bank: learnable
logits over $L$ candidate placements are mean-centered and transformed to a sparse
simplex-constrained weight vector $e \in \mathbb{R}^L$, inducing the support masks
$(\sigma_\uparrow,\, \sigma_\downarrow)$ that encode the structural support over directed
edges under the inferred motif mixture.
Magnitude is encoded separately through nonnegative strength matrices $\alpha$ and $\beta$,
enforced non-negative via softplus applied to unconstrained raw parameters; their Hadamard
products with the topology masks yield the effective interaction strengths
$\alpha^\mathrm{eff} = \alpha \odot \sigma_\uparrow$ and
$\beta^\mathrm{eff} = \beta \odot \sigma_\downarrow$ that enter the dynamics,
where $\odot$ denotes element-wise multiplication.
This factorization separates topology from magnitude estimation: FDU attention resolves
structural support through a sparse selection over the motif dictionary, while $(\alpha,
\beta)$ scale interaction strengths continuously from data.
Two implementation choices support stable operation over the large FDU bank: logits are
mean-centered before transformation, exploiting the shift-invariance of the entmax operator
for numerical stability; and the weighted sum over FDU masks is evaluated over the active
entmax support (entries where the attention weight is strictly positive), keeping bank
traversal tractable as $L$ scales with network size.
The entmax operator reduces to softmax when its concentration parameter equals one, enabling
smooth initialization; motif attention logits are initialized to zero, placing a uniform
prior over all $L$ candidate placements at the start of training so that the sparse
structural support emerges entirely through optimization with no initial preference among
FDUs.
The schedule governing concentration across training is described in the optimization
subsection.
The Hill parameters $\kappa$, $K$ and decay rates $\gamma$ are fixed rather than learned,
restricting inference to the interaction structure encoded in $(\alpha^\mathrm{eff},
\beta^\mathrm{eff})$.

A latent forcing module introduces a condition-dependent signal $w(t,u)$ that absorbs
contributions to the local dynamics that the FDU-regularized interaction structure cannot
represent, including influences from nodes outside the modeled subnetwork.
The forcing signal is expressed as $w(t,u) = V\,C(u)\,\Phi(t)$, where
$\Phi(t) \in \mathbb{R}^{1+2J}$ is a deterministic Fourier time basis evaluated
analytically at each query time, $C(u)$ is a single-hidden-layer network with hidden width 32 and $\tanh$
activation mapping the perturbation vector $u$ to a matrix of basis coefficients of shape
$r \times (1+2J)$, and
$V \in \mathbb{R}^{N \times r}$ is a learnable node mixing matrix.
Using an analytic Fourier basis with $J$ frequency components at time scale $\tau$ anchors
$w(t,u)$ to a smooth, finite-dimensional function space without learned temporal
parameters; condition dependence is encoded entirely through $C(u)$, so each perturbation
condition drives a distinct environmental trajectory while sharing the same temporal basis
structure.
The rank $r$ of $V$ controls the intrinsic complexity of the environmental drive: a low
rank reflects the assumption that a small number of latent programs account for
contributions from outside the modeled subnetwork.
Numerical values for all architectural parameters are specified in the Simulation and
Experimental Setup subsection.

\subsection{Training Objective}

The training objective is applied to the output of a differentiable chain: attention
weights over the FDU bank induce structural support masks that gate the interaction
strengths, which determine the ODE right-hand side against which the physics residual
is evaluated.
The composite loss differentiates through this entire chain, driving motif attention
toward the sparse FDU support consistent with the observed perturbation responses.
The resulting objective encodes a three-way tension between ODE consistency, data
fidelity, and structural regularization, with terms operating at different levels of
the inference problem.
The central term is the physics residual, which enforces that the predicted trajectory
satisfies the governing ODE at every training time point:
\begin{equation}
  \mathcal{L}_\mathrm{physics} = \left\|\frac{d\hat{x}}{dt}
    - \Bigl(F(\hat{x}) - \gamma\hat{x} + u + w(t,u)\Bigr)\right\|^2,
\end{equation}
where $F(\hat{x})$ is the net signed interaction drive computed from the Hill-type FDU
parameterization ($\alpha^\mathrm{eff}$, $\beta^\mathrm{eff}$, $\kappa$, $K$), $\gamma\hat{x}$
is the first-order decay term, $u$ is the perturbation input, and $w(t,u)$ is the latent
forcing signal.
Here $\|\cdot\|^2$ denotes the mean squared residual across the training batch and all $N$
state dimensions.

Four additional terms complete the training objective.
Data fidelity ($\mathcal{L}_\mathrm{data}$) is enforced as a point-wise mean squared error
between the predicted state and the observed trajectory, ensuring agreement with
ground-truth measurements at each queried time point.
Initial-condition anchoring ($\mathcal{L}_\mathrm{init}$) is enforced through a separate
forward pass on a dedicated batch of initial-time samples, giving explicit control over the
initial state independently of the physics residual.
Non-negativity ($\mathcal{L}_\mathrm{neg}$) is encouraged through a soft quadratic penalty
on the negative part of the predicted state, ensuring consistency with the nonnegative
domain required by the Hill-type interaction model.
Channel exclusivity ($\mathcal{L}_\mathrm{actinh}$) penalizes the mean of the elementwise
product $\alpha^\mathrm{eff} \odot \beta^\mathrm{eff}$, discouraging simultaneous activating
and inhibiting influence on the same directed edge; the penalty operates on the post-mask effective
strengths, so that only structurally supported interactions are subject to exclusivity
pressure.

These five terms are combined through an uncertainty-based adaptive weighting scheme
\textbf{\cite{Kendall2018}} in which each term carries a learnable log-variance parameter
$s_c$.
The uncertainty-weighted contribution of term $c$ is:
\begin{equation}
  \mathcal{L}_c^\mathrm{uw} = \tfrac{1}{2}\!\left(e^{-s_c}\,\mathcal{L}_c + s_c\right),
\end{equation}
where $e^{-s_c}$ is the effective weight and the additive $s_c$ prevents trivial solutions
in which all variances grow without bound.
The total loss is:
\begin{equation}
  \mathcal{L}_\mathrm{total} = \sum_c \mathcal{L}_c^\mathrm{uw}
    + \sum_k \lambda_k\,\mathcal{L}_k^\mathrm{forcing},
\end{equation}
where the first sum is over the five adaptive terms and the second over the forcing
penalties described below.
Log-variance parameters are initialized to establish a learning hierarchy:
$s_\mathrm{physics} = s_\mathrm{data} = -2$ (effective weight $e^2 \approx 7.4$), while
$s_\mathrm{init} = s_\mathrm{neg} = s_\mathrm{actinh} = 0$ (effective weight $1$).
This initialization ensures ODE consistency and trajectory fidelity dominate early in
training, preventing convergence to solutions that satisfy the structural priors before the
dynamics are correctly established.

The five adaptive terms correspond to objectives whose relative importance is genuinely
uncertain and data-dependent; the uncertainty-weighting mechanism allows their contributions
to adjust without manual tuning.
Four penalties governing the latent forcing signal are instead applied with pre-specified
weights $\lambda_k$ outside the adaptive tier.
These penalties are not imposed as first-principles laws of the environmental drive; they
function as identifiability-oriented regularizers that encode the intended role of $w(t,u)$
as a minimal, smooth residual with sparse node support that does not duplicate the
perturbation channel: an energy penalty ($\mathcal{L}_\mathrm{energy}$) controls the
overall magnitude of $w(t,u)$; a temporal smoothness penalty
($\mathcal{L}_\mathrm{smooth}$), computed via the same Jacobian-vector product mechanism,
enforces that $w(t,u)$ varies smoothly in time; a row-wise L2 penalty on $V$
($\mathcal{L}_\mathrm{row}$) encodes an identifiability prior, restricting exogenous
drive to a subset of nodes to prevent confounding of external influence with endogenous
regulation; and an anti-alignment penalty ($\mathcal{L}_\mathrm{align}$) enforces a
separation-of-roles constraint: $u$ is the perturbation channel carrying discriminative
structural information, and $w(t,u)$ must not reproduce the perturbation signal.
These penalties are placed outside the adaptive tier because the adaptive mechanism could
legitimately relax structural regularizers when the data warrants it, but unconstrained
forcing growth is not a valid optimization, since it would silently absorb structural signal
while appearing correct to the loss.

\subsection{Optimization Protocol}

The trajectory predictor, structural parameterizer, and latent forcing module are optimized
jointly using a single Adam optimizer with a shared learning rate; the forcing penalties
and structural regularizers govern how the residual is partitioned between them.

The concentration parameter of the entmax operator is annealed from the softmax limit,
placing a uniform prior over the FDU bank, to a near-sparsemax regime via a cosine
schedule: slow initially to allow broad exploration of the structural hypothesis space,
accelerating as the perturbation-response signal disambiguates the correct FDU support,
and decelerating again as concentration approaches its target to ease commitment
rather than impose it abruptly.

Log-variance parameters are clamped to a symmetric interval after each gradient step,
bounding the effective loss weights and preventing any single term from dominating or
vanishing entirely.
A learning rate scheduler reduces the learning rate when validation loss improvement falls
below a relative threshold over consecutive epochs, providing finer convergence after the
initial learning phase.
Gradient clipping bounds the gradient norm per update step, stabilizing the backward pass
through the physics residual.
Training terminates when validation loss shows no improvement for a fixed patience window;
the model checkpoint from the best-performing validation epoch is restored at convergence.

\subsection{Simulation and Experimental Setup}

Ground-truth trajectories are generated by numerical integration of the FDU-regularized ODE
with known interaction structure using the RK45 (Runge-Kutta 45) solver over $t \in [0,\, 100]$.
All node states are initialized uniformly to $x_i(0) = 5$.
Hill function parameters are set to $\kappa = 2$ and $K = 20$, uniform across all source
nodes, and first-order decay rates to $\gamma = 0.1$ for all nodes.
Ground-truth interaction strengths are binary: each directed edge in the FDU-constructed
network carries unit strength ($\alpha_{ij} = 1$ or $\beta_{ij} = 1$).
With $\kappa = 2$, the Hill half-activation threshold is $K^{1/\kappa} = \sqrt{20}
\approx 4.5$; the uniform initial state $x_i(0) = 5$ lies near this threshold, placing the
system in the sensitive dynamical regime where source activity changes produce the largest
fractional response and perturbation-driven deviations carry maximal structural information.

The 20-node ground-truth network was constructed by placing 10 signed triad subnetworks,
drawn uniformly at random from the 512 admissible signed triads, onto randomly selected
node triplets under a fixed random seed.
Construction enforces three per-node sparsity constraints: signed in-degree and out-degree
are each bounded at 3, each node participates in at most 2 embedded triads, and no directed
edge is reused across triads.
The resulting network comprises 43 directed signed edges spanning 15 unidirectional node
pairs and 14 mutual node pairs.

Observations are sampled sparsely across four temporally disjoint windows designed to
capture distinct dynamical phases: initialization ($t \in [0,\, 0.1]$), early-to-mid
transient ($t \in [1,\, 30]$), mid-to-late transient ($t \in [40,\, 60]$), and late steady
state ($t \in [70,\, 90]$).
Spanning these phases distributes structural information across the full trajectory, and
the boundaries are fixed across all experiments rather than tuned to individual motif
trajectories.
Each window contributes 75 training and 25 validation time points, drawn randomly from a
uniform distribution within the window.
A held-out test set of 100 time points is drawn uniformly from the full horizon
$t \in [0,\, 100]$, spanning both within-window and gap regions; approximately $31\%$ of
the horizon lies outside any training window, providing evaluation in temporally unobserved
regions.

Perturbation panels are assigned per the structural class of the target motif.
Each condition is specified by a signed amplitude vector $u \in \mathbb{R}^N$: non-zero
entries carry a signed amplitude of fixed magnitude $1.0$, uniform across nodes and equal
in absolute value for positive and negative directions, and all other entries are zero;
the unperturbed reference sets $u = \mathbf{0}$.
Each non-zero $u_i$ is applied as a smoothly gated step with sigmoidal onset at $t = 0$ and
deactivation at the end of the simulation horizon, with a transition timescale of 1 time unit;
the sigmoidal profile avoids discontinuities in the ODE right-hand side.
Single-component conditions perturb one node at a time in both directions; co-perturbation
conditions perturb two nodes simultaneously with same-sign or opposite-sign combinations as
specified by the structural class.
Conditions cover all nodes and node pairs in the target subset exhaustively.

The trajectory predictor uses depth 6 and hidden width 512, with layer normalization and
dropout rate $0.1$.
The entmax concentration is annealed from $1.0$ to $1.5$ via a cosine schedule:
\begin{equation}
  c(\delta) = c_i + (c_f - c_i) \cdot \frac{1 - \cos(\pi \delta / T_{\mathrm{max}})}{2},
\end{equation}
where $c_i$ and $c_f$ are the initial and final concentration values, $\delta$ is the
current epoch, and $T_{\mathrm{max}}$ the total number of epochs.
The latent forcing module, active in the 20-node experiment, uses $J = 4$ Fourier frequency
components, time scale $\tau = 100$, and rank $r = 1$; the latent forcing module is disabled
in the triad-only experiments, where the closed-system assumption renders environmental
contributions structurally absent.
Forcing penalty weights follow the same cosine schedule form as the entmax concentration,
decaying from $10^{-3}$ to $10^{-4}$ over training for all four penalties
($\mathcal{L}_\mathrm{energy}$, $\mathcal{L}_\mathrm{smooth}$,
$\mathcal{L}_\mathrm{row}$, $\mathcal{L}_\mathrm{align}$); the decay reflects that strong early
regularization prevents forcing from absorbing structural signal during the critical early
phase, while relaxing constraints later allows it to absorb genuine residuals.
All parameters are optimized with Adam at learning rate $1 \times 10^{-4}$ with batch size
50, for a maximum of 1000 epochs.
The learning rate scheduler applies a reduction factor of $0.5$ when validation loss
improvement falls below $10^{-3}$ relative over 10 consecutive epochs, with a minimum
learning rate of $10^{-6}$.
Gradient norms are clipped to a maximum of $5.0$ per update step.
Log-variance parameters are clamped within $[-\log(150),\, \log(150)]$ after each gradient
step, bounding effective loss weights to the interval $[1/150,\, 150]$.
Early stopping is applied with a patience of 50 epochs.

\subsection{Computational Resources}

All experiments were run on a CPU-only platform (2.50 GHz, 128 GiB RAM).
Each 3-node experiment required approximately $2.74$ s per epoch (wall-clock, including
training, validation, and checkpointing), corresponding to approximately 46 min for 1000
epochs.
The 20-node experiment required approximately $5.62$ s per epoch, corresponding to
approximately 1 h 34 min.
Although the 20-node FDU bank is approximately 1000-fold larger than the 3-node bank
(73,000 vs 70 candidate placements), training time increased by less than twofold, a
direct consequence of entmax sparsification concentrating computation onto a small active
support regardless of bank size.
Experiments were implemented in Python 3.12 using PyTorch 2.10 for model training and
automatic differentiation, SciPy 1.17 (RK45 solver) for ground-truth ODE integration,
and NumPy 1.26 for numerical operations.


\section*{Data Availability}

All datasets used in this study are synthetic and were generated computationally using
the code available at \url{https://github.com/AstraZeneca/fdu-pisi}.
No empirical data were collected or used.
The datasets generated and/or analysed during the current study are available in the
Zenodo repository, \url{https://doi.org/10.5281/zenodo.20377559}.

\section*{Code Availability}

The source code, documentation, and all materials required to reproduce the experiments
are publicly available on GitHub at \url{https://github.com/AstraZeneca/fdu-pisi}.


\section*{Acknowledgments}

The author is grateful to Dr. Virginia Savova for supervision, strategic direction, and critical reading of the manuscript, and AstraZeneca Research and Development for institutional support.

\section*{Funding}

This work was supported by AstraZeneca US.

\section*{Author Contributions}

N.N. conducted all research and authored the paper.

\section*{Competing Interests}

The author is an employee of AstraZeneca US.

{\small
\setlength{\bibsep}{0pt plus 0.3ex}
\bibliography{references}

\begin{thebibliography}{10}
\expandafter\ifx\csname url\endcsname\relax
  \def\url#1{\texttt{#1}}\fi
\expandafter\ifx\csname urlprefix\endcsname\relax\def\urlprefix{URL }\fi
\providecommand{\bibinfo}[2]{#2}
\providecommand{\eprint}[2][]{\url{#2}}

\bibitem{Pearl2009}
\bibinfo{author}{Pearl, J.}
\newblock \emph{\bibinfo{title}{Causality: Models, Reasoning, and Inference}}
  (\bibinfo{publisher}{Cambridge University Press}, \bibinfo{year}{2009}),
  \bibinfo{edition}{2nd} edn.

\bibitem{Spirtes2000}
\bibinfo{author}{Spirtes, P.}, \bibinfo{author}{Glymour, C.~N.} \&
  \bibinfo{author}{Scheines, R.}
\newblock \emph{\bibinfo{title}{Causation, Prediction, and Search}}
  (\bibinfo{publisher}{MIT Press}, \bibinfo{year}{2000}).

\bibitem{Hauser2012}
\bibinfo{author}{Hauser, A.} \& \bibinfo{author}{B{\"u}hlmann, P.}
\newblock \bibinfo{title}{Characterization and greedy learning of
  interventional markov equivalence classes of directed acyclic graphs}.
\newblock \emph{\bibinfo{journal}{The Journal of Machine Learning Research}}
  \textbf{\bibinfo{volume}{13}}, \bibinfo{pages}{2409--2464}
  (\bibinfo{year}{2012}).

\bibitem{Aalto2020}
\bibinfo{author}{Aalto, A.}, \bibinfo{author}{Viitasaari, L.},
  \bibinfo{author}{Ilmonen, P.}, \bibinfo{author}{Mombaerts, L.} \&
  \bibinfo{author}{Gon{\c{c}}alves, J.}
\newblock \bibinfo{title}{Gene regulatory network inference from sparsely
  sampled noisy data}.
\newblock \emph{\bibinfo{journal}{Nature Communications}}
  \textbf{\bibinfo{volume}{11}}, \bibinfo{pages}{3493} (\bibinfo{year}{2020}).

\bibitem{Sarmah2022}
\bibinfo{author}{Sarmah, D.} \emph{et~al.}
\newblock \bibinfo{title}{Network inference from perturbation time course
  data}.
\newblock \emph{\bibinfo{journal}{npj Systems Biology and Applications}}
  \textbf{\bibinfo{volume}{8}}, \bibinfo{pages}{42} (\bibinfo{year}{2022}).

\bibitem{Runge2023}
\bibinfo{author}{Runge, J.}, \bibinfo{author}{Gerhardus, A.},
  \bibinfo{author}{Varando, G.}, \bibinfo{author}{Eyring, V.} \&
  \bibinfo{author}{Camps-Valls, G.}
\newblock \bibinfo{title}{Causal inference for time series}.
\newblock \emph{\bibinfo{journal}{Nature Reviews Earth \& Environment}}
  \textbf{\bibinfo{volume}{4}}, \bibinfo{pages}{487--505}
  (\bibinfo{year}{2023}).

\bibitem{HeGeng2008}
\bibinfo{author}{He, Y.-B.} \& \bibinfo{author}{Geng, Z.}
\newblock \bibinfo{title}{Active learning of causal networks with intervention
  experiments and optimal designs}.
\newblock \emph{\bibinfo{journal}{Journal of Machine Learning Research}}
  \textbf{\bibinfo{volume}{9}}, \bibinfo{pages}{2523--2547}
  (\bibinfo{year}{2008}).

\bibitem{Gutenkunst2007}
\bibinfo{author}{Gutenkunst, R.~N.} \emph{et~al.}
\newblock \bibinfo{title}{Universally sloppy parameter sensitivities in systems
  biology models}.
\newblock \emph{\bibinfo{journal}{PLoS Computational Biology}}
  \textbf{\bibinfo{volume}{3}}, \bibinfo{pages}{e189} (\bibinfo{year}{2007}).

\bibitem{Polynikis2009}
\bibinfo{author}{Polynikis, A.}, \bibinfo{author}{Hogan, S.~J.} \&
  \bibinfo{author}{Di~Bernardo, M.}
\newblock \bibinfo{title}{Comparing different {ODE} modelling approaches for
  gene regulatory networks}.
\newblock \emph{\bibinfo{journal}{Journal of Theoretical Biology}}
  \textbf{\bibinfo{volume}{261}}, \bibinfo{pages}{511--530}
  (\bibinfo{year}{2009}).

\bibitem{Scholkopf2021}
\bibinfo{author}{Sch{\"o}lkopf, B.} \emph{et~al.}
\newblock \bibinfo{title}{Toward causal representation learning}.
\newblock \emph{\bibinfo{journal}{Proceedings of the IEEE}}
  \textbf{\bibinfo{volume}{109}}, \bibinfo{pages}{612--634}
  (\bibinfo{year}{2021}).

\bibitem{PetersBook2017}
\bibinfo{author}{Peters, J.}, \bibinfo{author}{Janzing, D.} \&
  \bibinfo{author}{Sch{\"o}lkopf, B.}
\newblock \emph{\bibinfo{title}{Elements of Causal Inference: Foundations and
  Learning Algorithms}} (\bibinfo{publisher}{MIT Press}, \bibinfo{year}{2017}).

\bibitem{EdwardsGlass2000}
\bibinfo{author}{Edwards, R.} \& \bibinfo{author}{Glass, L.}
\newblock \bibinfo{title}{Combinatorial explosion in model gene networks}.
\newblock \emph{\bibinfo{journal}{Chaos: An Interdisciplinary Journal of
  Nonlinear Science}} \textbf{\bibinfo{volume}{10}}, \bibinfo{pages}{691--704}
  (\bibinfo{year}{2000}).

\bibitem{Milo2002}
\bibinfo{author}{Milo, R.} \emph{et~al.}
\newblock \bibinfo{title}{Network motifs: Simple building blocks of complex
  networks}.
\newblock \emph{\bibinfo{journal}{Science}} \textbf{\bibinfo{volume}{298}},
  \bibinfo{pages}{824--827} (\bibinfo{year}{2002}).

\bibitem{Alon2007}
\bibinfo{author}{Alon, U.}
\newblock \bibinfo{title}{Network motifs: theory and experimental approaches}.
\newblock \emph{\bibinfo{journal}{Nature Reviews Genetics}}
  \textbf{\bibinfo{volume}{8}}, \bibinfo{pages}{450--461}
  (\bibinfo{year}{2007}).

\bibitem{ManganAlon2003}
\bibinfo{author}{Mangan, S.} \& \bibinfo{author}{Alon, U.}
\newblock \bibinfo{title}{Structure and function of the feed-forward loop
  network motif}.
\newblock \emph{\bibinfo{journal}{Proceedings of the National Academy of
  Sciences}} \textbf{\bibinfo{volume}{100}}, \bibinfo{pages}{11980--11985}
  (\bibinfo{year}{2003}).

\bibitem{Benson2016}
\bibinfo{author}{Benson, A.~R.}, \bibinfo{author}{Gleich, D.~F.} \&
  \bibinfo{author}{Leskovec, J.}
\newblock \bibinfo{title}{Higher-order organization of complex networks}.
\newblock \emph{\bibinfo{journal}{Science}} \textbf{\bibinfo{volume}{353}},
  \bibinfo{pages}{163--166} (\bibinfo{year}{2016}).

\bibitem{Eberhardt2007}
\bibinfo{author}{Eberhardt, F.} \& \bibinfo{author}{Scheines, R.}
\newblock \bibinfo{title}{Interventions and causal inference}.
\newblock \emph{\bibinfo{journal}{Philosophy of Science}}
  \textbf{\bibinfo{volume}{74}}, \bibinfo{pages}{981--995}
  (\bibinfo{year}{2007}).

\bibitem{Kent2013}
\bibinfo{author}{Kent, E.}, \bibinfo{author}{Neumann, S.},
  \bibinfo{author}{Kummer, U.} \& \bibinfo{author}{Mendes, P.}
\newblock \bibinfo{title}{What can we learn from global sensitivity analysis of
  biochemical systems?}
\newblock \emph{\bibinfo{journal}{PLoS ONE}} \textbf{\bibinfo{volume}{8}},
  \bibinfo{pages}{e79244} (\bibinfo{year}{2013}).

\bibitem{Stepaniants2020}
\bibinfo{author}{Stepaniants, G.}, \bibinfo{author}{Brunton, B.~W.} \&
  \bibinfo{author}{Kutz, J.~N.}
\newblock \bibinfo{title}{Inferring causal networks of dynamical systems
  through transient dynamics and perturbation}.
\newblock \emph{\bibinfo{journal}{Physical Review E}}
  \textbf{\bibinfo{volume}{102}}, \bibinfo{pages}{042309}
  (\bibinfo{year}{2020}).

\bibitem{Granger1969}
\bibinfo{author}{Granger, C. W.~J.}
\newblock \bibinfo{title}{Investigating causal relations by econometric models
  and cross-spectral methods}.
\newblock \emph{\bibinfo{journal}{Econometrica}} \textbf{\bibinfo{volume}{37}},
  \bibinfo{pages}{424--438} (\bibinfo{year}{1969}).

\bibitem{Huynh-Thu2018}
\bibinfo{author}{Huynh-Thu, V.~A.} \& \bibinfo{author}{Geurts, P.}
\newblock \bibinfo{title}{{dynGENIE3}: Dynamical {GENIE3} for the inference of
  gene networks from time series expression data}.
\newblock \emph{\bibinfo{journal}{Scientific Reports}}
  \textbf{\bibinfo{volume}{8}}, \bibinfo{pages}{3384} (\bibinfo{year}{2018}).

\bibitem{Brunton2016}
\bibinfo{author}{Brunton, S.~L.}, \bibinfo{author}{Proctor, J.~L.} \&
  \bibinfo{author}{Kutz, J.~N.}
\newblock \bibinfo{title}{Discovering governing equations from data by sparse
  identification of nonlinear dynamical systems}.
\newblock \emph{\bibinfo{journal}{Proceedings of the National Academy of
  Sciences}} \textbf{\bibinfo{volume}{113}}, \bibinfo{pages}{3932--3937}
  (\bibinfo{year}{2016}).

\bibitem{Timme2014}
\bibinfo{author}{Timme, M.} \& \bibinfo{author}{Casadiego, J.}
\newblock \bibinfo{title}{Revealing networks from dynamics: An introduction}.
\newblock \emph{\bibinfo{journal}{Journal of Physics A: Mathematical and
  Theoretical}} \textbf{\bibinfo{volume}{47}}, \bibinfo{pages}{343001}
  (\bibinfo{year}{2014}).

\bibitem{Pamfil2020}
\bibinfo{author}{Pamfil, R.} \emph{et~al.}
\newblock \bibinfo{title}{{DYNOTEARS}: Structure learning from time-series
  data}.
\newblock In \emph{\bibinfo{booktitle}{International Conference on Artificial
  Intelligence and Statistics}}, \bibinfo{pages}{1595--1605}
  (\bibinfo{organization}{PMLR}, \bibinfo{year}{2020}).

\bibitem{Kipf2018}
\bibinfo{author}{Kipf, T.}, \bibinfo{author}{Fetaya, E.},
  \bibinfo{author}{Wang, K.-C.}, \bibinfo{author}{Welling, M.} \&
  \bibinfo{author}{Zemel, R.}
\newblock \bibinfo{title}{Neural relational inference for interacting systems}.
\newblock In \emph{\bibinfo{booktitle}{International Conference on Machine
  Learning}}, \bibinfo{pages}{2688--2697} (\bibinfo{organization}{PMLR},
  \bibinfo{year}{2018}).

\bibitem{Huang2020graphode}
\bibinfo{author}{Huang, Z.}, \bibinfo{author}{Sun, Y.} \&
  \bibinfo{author}{Wang, W.}
\newblock \bibinfo{title}{Learning continuous system dynamics from
  irregularly-sampled partial observations}.
\newblock \emph{\bibinfo{journal}{Advances in Neural Information Processing
  Systems}} \textbf{\bibinfo{volume}{33}}, \bibinfo{pages}{16177--16187}
  (\bibinfo{year}{2020}).

\bibitem{Bhaskar2024}
\bibinfo{author}{Bhaskar, D.} \emph{et~al.}
\newblock \bibinfo{title}{Inferring dynamic regulatory interaction graphs from
  time series data with perturbations}.
\newblock In \emph{\bibinfo{booktitle}{Proceedings of the Second Learning on
  Graphs Conference}}, vol. \bibinfo{volume}{231} of
  \emph{\bibinfo{series}{Proceedings of Machine Learning Research}},
  \bibinfo{pages}{22:1--22:21} (\bibinfo{publisher}{PMLR},
  \bibinfo{year}{2024}).

\bibitem{Peters2013timeseries}
\bibinfo{author}{Peters, J.}, \bibinfo{author}{Janzing, D.} \&
  \bibinfo{author}{Sch{\"o}lkopf, B.}
\newblock \bibinfo{title}{Causal inference on time series using restricted
  structural equation models}.
\newblock \emph{\bibinfo{journal}{Advances in Neural Information Processing
  Systems}} \textbf{\bibinfo{volume}{26}} (\bibinfo{year}{2013}).

\bibitem{Runge2019}
\bibinfo{author}{Runge, J.}, \bibinfo{author}{Nowack, P.},
  \bibinfo{author}{Kretschmer, M.}, \bibinfo{author}{Flaxman, S.} \&
  \bibinfo{author}{Sejdinovic, D.}
\newblock \bibinfo{title}{Detecting and quantifying causal associations in
  large nonlinear time series datasets}.
\newblock \emph{\bibinfo{journal}{Science Advances}}
  \textbf{\bibinfo{volume}{5}}, \bibinfo{pages}{eaau4996}
  (\bibinfo{year}{2019}).

\bibitem{Chen2018}
\bibinfo{author}{Chen, R. T.~Q.}, \bibinfo{author}{Rubanova, Y.},
  \bibinfo{author}{Bettencourt, J.} \& \bibinfo{author}{Duvenaud, D.}
\newblock \bibinfo{title}{Neural ordinary differential equations}.
\newblock In \emph{\bibinfo{booktitle}{Advances in Neural Information
  Processing Systems}}, vol.~\bibinfo{volume}{31} (\bibinfo{year}{2018}).

\bibitem{Kidger2020}
\bibinfo{author}{Kidger, P.}, \bibinfo{author}{Morrill, J.},
  \bibinfo{author}{Foster, J.} \& \bibinfo{author}{Lyons, T.}
\newblock \bibinfo{title}{Neural controlled differential equations for
  irregular time series}.
\newblock \emph{\bibinfo{journal}{Advances in Neural Information Processing
  Systems}} \textbf{\bibinfo{volume}{33}}, \bibinfo{pages}{6696--6707}
  (\bibinfo{year}{2020}).

\bibitem{Raissi2019}
\bibinfo{author}{Raissi, M.}, \bibinfo{author}{Perdikaris, P.} \&
  \bibinfo{author}{Karniadakis, G.~E.}
\newblock \bibinfo{title}{Physics-informed neural networks: A deep learning
  framework for solving forward and inverse problems involving nonlinear
  partial differential equations}.
\newblock \emph{\bibinfo{journal}{Journal of Computational Physics}}
  \textbf{\bibinfo{volume}{378}}, \bibinfo{pages}{686--707}
  (\bibinfo{year}{2019}).

\bibitem{Karniadakis2021}
\bibinfo{author}{Karniadakis, G.~E.} \emph{et~al.}
\newblock \bibinfo{title}{Physics-informed machine learning}.
\newblock \emph{\bibinfo{journal}{Nature Reviews Physics}}
  \textbf{\bibinfo{volume}{3}}, \bibinfo{pages}{422--440}
  (\bibinfo{year}{2021}).

\bibitem{YuWang2024}
\bibinfo{author}{Yu, R.} \& \bibinfo{author}{Wang, R.}
\newblock \bibinfo{title}{Learning dynamical systems from data: An introduction
  to physics-guided deep learning}.
\newblock \emph{\bibinfo{journal}{Proceedings of the National Academy of
  Sciences}} \textbf{\bibinfo{volume}{121}}, \bibinfo{pages}{e2311808121}
  (\bibinfo{year}{2024}).

\bibitem{Mircea2024}
\bibinfo{author}{Mircea, M.}, \bibinfo{author}{Garlaschelli, D.} \&
  \bibinfo{author}{Semrau, S.}
\newblock \bibinfo{title}{Inference of dynamical gene regulatory networks from
  single-cell data with physics informed neural networks}.
\newblock \emph{\bibinfo{journal}{arXiv preprint arXiv:2401.07379}}
  (\bibinfo{year}{2024}).

\bibitem{Peters2019entmax}
\bibinfo{author}{Peters, B.}, \bibinfo{author}{Niculae, V.} \&
  \bibinfo{author}{Martins, A. F.~T.}
\newblock \bibinfo{title}{Sparse sequence-to-sequence models}.
\newblock In \emph{\bibinfo{booktitle}{Proceedings of the 57th Annual Meeting
  of the Association for Computational Linguistics}},
  \bibinfo{pages}{1504--1519} (\bibinfo{year}{2019}).

\bibitem{Hill1910}
\bibinfo{author}{Hill, A.~V.}
\newblock \bibinfo{title}{The possible effects of the aggregation of the
  molecules of haemoglobin on its dissociation curves}.
\newblock \emph{\bibinfo{journal}{The Journal of Physiology}}
  \textbf{\bibinfo{volume}{40}}, \bibinfo{pages}{iv--vii}
  (\bibinfo{year}{1910}).

\bibitem{Alvarez2009latentforce}
\bibinfo{author}{Alvarez, M.}, \bibinfo{author}{Luengo, D.} \&
  \bibinfo{author}{Lawrence, N.~D.}
\newblock \bibinfo{title}{Latent force models}.
\newblock In \emph{\bibinfo{booktitle}{Artificial Intelligence and
  Statistics}}, \bibinfo{pages}{9--16} (\bibinfo{organization}{PMLR},
  \bibinfo{year}{2009}).

\bibitem{Rackauckas2020ude}
\bibinfo{author}{Rackauckas, C.} \emph{et~al.}
\newblock \bibinfo{title}{Universal differential equations for scientific
  machine learning}.
\newblock \emph{\bibinfo{journal}{arXiv preprint arXiv:2001.04385}}
  (\bibinfo{year}{2020}).

\bibitem{Lin2025PerturbODE}
\bibinfo{author}{Lin, Z.} \emph{et~al.}
\newblock \bibinfo{title}{Interpretable neural {ODEs} for gene regulatory
  network discovery under perturbations}.
\newblock \emph{\bibinfo{journal}{arXiv preprint arXiv:2501.02409}}
  (\bibinfo{year}{2025}).
\newblock \bibinfo{note}{Preprint}.

\bibitem{de2002modeling}
\bibinfo{author}{De~Jong, H.}
\newblock \bibinfo{title}{Modeling and simulation of genetic regulatory
  systems: A literature review}.
\newblock \emph{\bibinfo{journal}{Journal of Computational Biology}}
  \textbf{\bibinfo{volume}{9}}, \bibinfo{pages}{67--103}
  (\bibinfo{year}{2002}).

\bibitem{lee2002transcriptional}
\bibinfo{author}{Lee, T.~I.} \emph{et~al.}
\newblock \bibinfo{title}{Transcriptional regulatory networks in {Saccharomyces
  cerevisiae}}.
\newblock \emph{\bibinfo{journal}{Science}} \textbf{\bibinfo{volume}{298}},
  \bibinfo{pages}{799--804} (\bibinfo{year}{2002}).

\bibitem{Leskovec2010signed}
\bibinfo{author}{Leskovec, J.}, \bibinfo{author}{Huttenlocher, D.} \&
  \bibinfo{author}{Kleinberg, J.}
\newblock \bibinfo{title}{Signed networks in social media}.
\newblock In \emph{\bibinfo{booktitle}{Proceedings of the SIGCHI Conference on
  Human Factors in Computing Systems}}, \bibinfo{pages}{1361--1370}
  (\bibinfo{year}{2010}).

\bibitem{InoueTodo2019}
\bibinfo{author}{Inoue, H.} \& \bibinfo{author}{Todo, Y.}
\newblock \bibinfo{title}{Firm-level propagation of shocks through supply-chain
  networks}.
\newblock \emph{\bibinfo{journal}{Nature Sustainability}}
  \textbf{\bibinfo{volume}{2}}, \bibinfo{pages}{841--847}
  (\bibinfo{year}{2019}).

\bibitem{Acemoglu2012network}
\bibinfo{author}{Acemoglu, D.}, \bibinfo{author}{Carvalho, V.~M.},
  \bibinfo{author}{Ozdaglar, A.} \& \bibinfo{author}{Tahbaz-Salehi, A.}
\newblock \bibinfo{title}{The network origins of aggregate fluctuations}.
\newblock \emph{\bibinfo{journal}{Econometrica}} \textbf{\bibinfo{volume}{80}},
  \bibinfo{pages}{1977--2016} (\bibinfo{year}{2012}).

\bibitem{Ba2016}
\bibinfo{author}{Ba, J.~L.}, \bibinfo{author}{Kiros, J.~R.} \&
  \bibinfo{author}{Hinton, G.~E.}
\newblock \bibinfo{title}{Layer normalization}.
\newblock \emph{\bibinfo{journal}{arXiv preprint arXiv:1607.06450}}
  (\bibinfo{year}{2016}).

\bibitem{Srivastava2014}
\bibinfo{author}{Srivastava, N.}, \bibinfo{author}{Hinton, G.},
  \bibinfo{author}{Krizhevsky, A.}, \bibinfo{author}{Sutskever, I.} \&
  \bibinfo{author}{Salakhutdinov, R.}
\newblock \bibinfo{title}{Dropout: A simple way to prevent neural networks from
  overfitting}.
\newblock \emph{\bibinfo{journal}{Journal of Machine Learning Research}}
  \textbf{\bibinfo{volume}{15}}, \bibinfo{pages}{1929--1958}
  (\bibinfo{year}{2014}).

\bibitem{Glorot2010}
\bibinfo{author}{Glorot, X.} \& \bibinfo{author}{Bengio, Y.}
\newblock \bibinfo{title}{Understanding the difficulty of training deep
  feedforward neural networks}.
\newblock In \emph{\bibinfo{booktitle}{Proceedings of the Thirteenth
  International Conference on Artificial Intelligence and Statistics}},
  \bibinfo{pages}{249--256} (\bibinfo{organization}{JMLR Workshop and
  Conference Proceedings}, \bibinfo{year}{2010}).

\bibitem{Kendall2018}
\bibinfo{author}{Kendall, A.}, \bibinfo{author}{Gal, Y.} \&
  \bibinfo{author}{Cipolla, R.}
\newblock \bibinfo{title}{Multi-task learning using uncertainty to weigh losses
  for scene geometry and semantics}.
\newblock In \emph{\bibinfo{booktitle}{Proceedings of the IEEE Conference on
  Computer Vision and Pattern Recognition}}, \bibinfo{pages}{7482--7491}
  (\bibinfo{year}{2018}).

\end{thebibliography}
}

\clearpage
\onecolumn

\clearpage
\noindent\includegraphics[width=\textwidth]{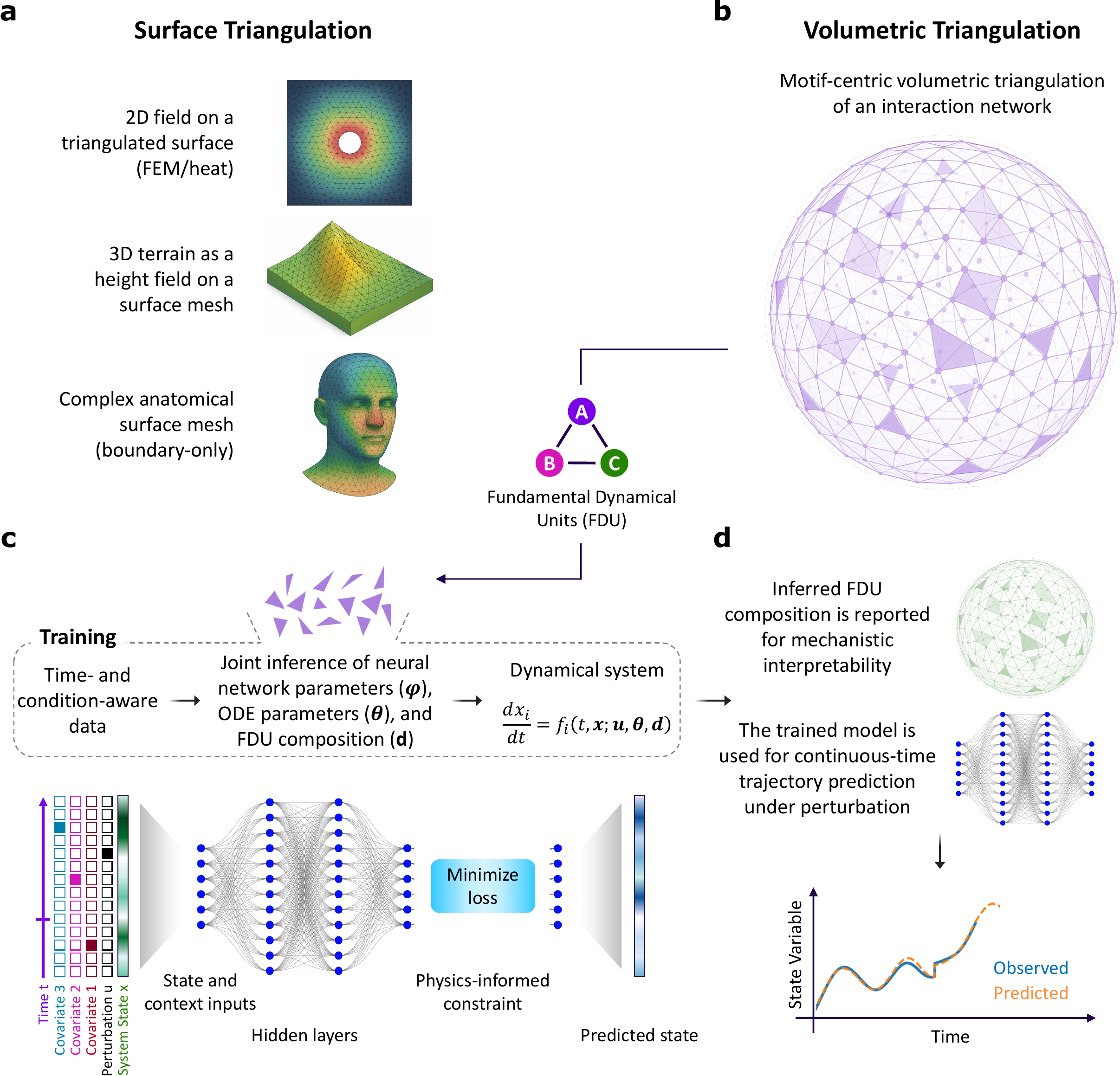}
\captionof{figure}{%
    \textbf{Motif-centric volumetric triangulation and FDU-regularized physics-informed
    neural ODE framework for joint trajectory prediction and structural inference.}
    \textbf{(a)}~Surface triangulation encodes geometry at the boundary: (top) a scalar
    field on a flat triangulated mesh (finite-element heat diffusion); (middle) a
    three-dimensional terrain represented as a height field on a surface mesh; (bottom) a
    complex surface mesh capturing only boundary topology.
    In each case, triangulation defines connectivity among surface elements without imposing
    constraints on the interior.
    In network terms, a conventional directed graph is the surface analog: pairwise
    connectivity encoded without constraining how edges compose into multi-path structures.
    \textbf{(b)}~Volumetric triangulation of an interaction network.
    Each Fundamental Dynamical Unit (FDU) is a signed three-node subgraph (inset: nodes A,
    B, C with signed directed edges) that couples direct and relayed interaction pathways
    within a single primitive.
    A network is represented as a sparse superposition of FDUs, visualised here as
    interior-filling triangular primitives within a spherical interaction network (purple),
    so that every inferred structure carries explicit constraints on interior pathway
    composition, not only on pairwise connectivity.
    \textbf{(c)}~Joint training procedure.
    Time- and condition-aware observations (time $t$, system state $x$, and perturbation
    input $\mathbf{u}$) are passed to a neural network that produces the predicted state
    and jointly infers network parameters $\boldsymbol{\phi}$, ODE parameters
    $\boldsymbol{\theta}$, and FDU composition $\mathbf{d}$.
    A physics-informed constraint layer penalises violations of the governing ODE during
    backpropagation, coupling data fidelity with dynamical consistency throughout training.
    \textbf{(d)}~Inference outputs of the trained model.
    Top: the inferred FDU composition is extracted as a sparse attention-weighted
    superposition over the FDU bank and reported as an interaction network (green) for
    mechanistic interpretability.
    Middle and Bottom: the trained model integrates the learned ODE forward in time to
    produce continuous-time trajectory predictions (orange, dashed) under arbitrary
    perturbation conditions; predicted trajectories are shown against observed state
    variables (blue, solid).
  }
\label{fig:1}
\vspace{1em}

\clearpage
\noindent\includegraphics[width=\textwidth]{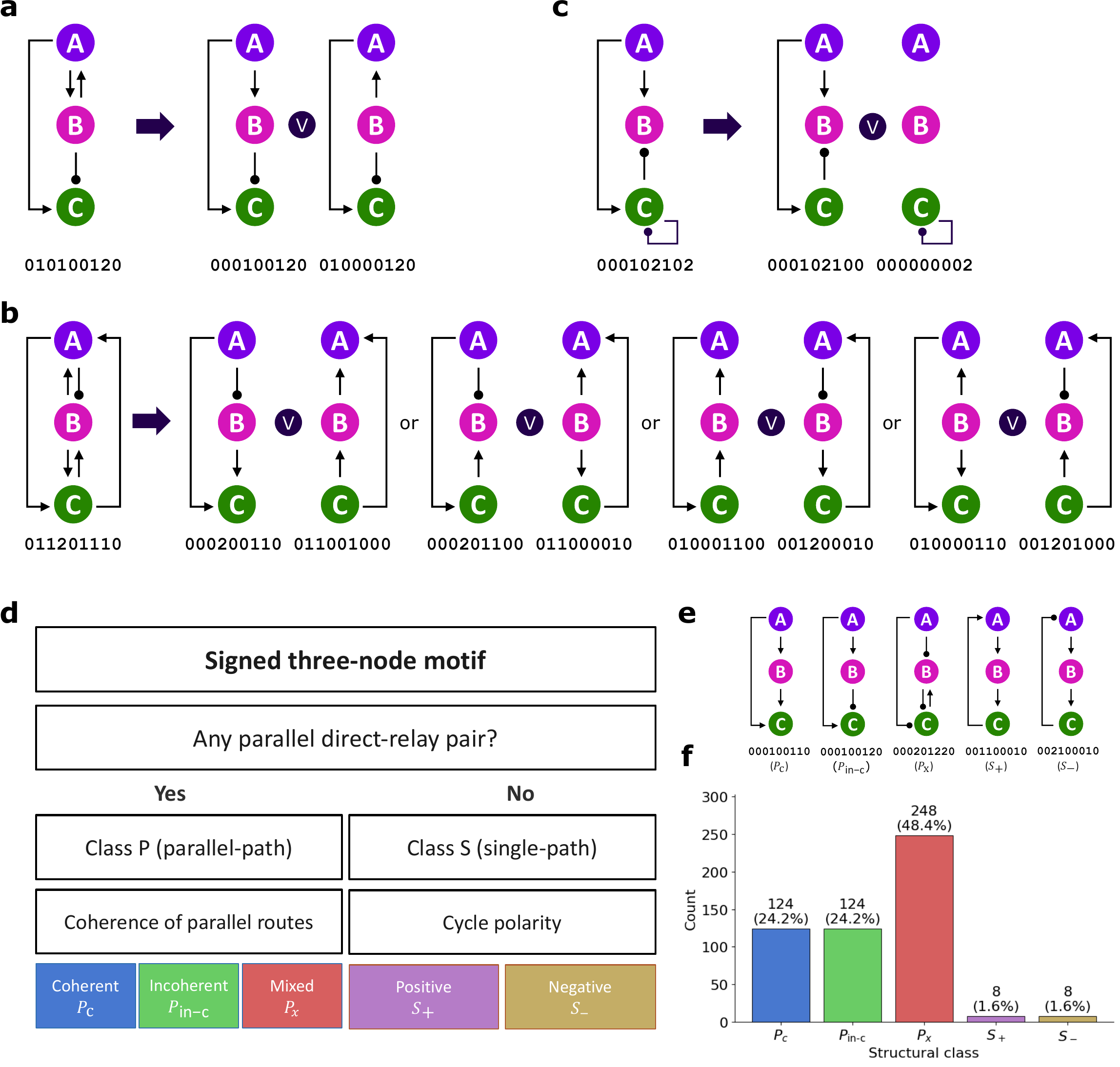}
\captionof{figure}{%
    \textbf{FDU dictionary construction and permutation-invariant structural classification
    of signed three-node triads.}
    \textbf{(a)}~FDU decomposition of a triad with one mutual pair ($m=1$).
    The pairwise-connected triad \texttt{010100120}, containing a mutual activating
    interaction on $\{A,B\}$, is not a tournament and admits a unique two-FDU
    reconstruction via superposition ($\vee$):
    \texttt{000100120} $\vee$ \texttt{010000120}.
    \textbf{(b)}~FDU decomposition of a triad with three mutual pairs ($m=3$).
    The triad \texttt{011201110} admits exactly $2^{m-1}=4$ distinct minimal two-FDU
    decompositions; each pair jointly realizes all three mutual interactions in opposite
    directions while preserving forced edge orientations and signs.
    \textbf{(c)}~Extension to self-regulation.
    The triad \texttt{000102102}, combining a triadic architecture with an auto-inhibiting
    self-edge on node C, is expressed as the superposition of the triadic FDU
    \texttt{000102100} and the unary auto-inhibiting self-motif \texttt{000000002}.
    \textbf{(d)}~Taxonomy decision tree.
    Each signed triad is tested for the presence of at least one parallel ordered pair
    (class $P$: parallel-path; class $S$: single-path), then refined by coherence state
    within class $P$ ($P_c$: coherent; $P_\mathrm{in\text{-}c}$: incoherent; $P_x$: mixed)
    or by cycle polarity within class $S$ ($S_+$: positive; $S_-$: negative).
    \textbf{(e)}~One canonical representative per structural class:
    \texttt{000100110} ($P_c$), \texttt{000100120} ($P_\mathrm{in\text{-}c}$),
    \texttt{000201220} ($P_x$), \texttt{001100010} ($S_+$), \texttt{002100010} ($S_-$).
    \textbf{(f)}~Distribution of all 512 pairwise-connected signed triads across the five
    structural classes; count and percentage of total shown above each bar.
    Bar colors correspond to the structural class color bands in panel~(d).
    In panels~a--c and~e, arrows denote activating edges, filled circles denote inhibiting
    edges, and nodes are labeled A, B, C (equivalently, nodes 1, 2, 3 in the main text).
  }
\label{fig:2}
\vspace{1em}

\clearpage
\noindent\includegraphics[width=\textwidth]{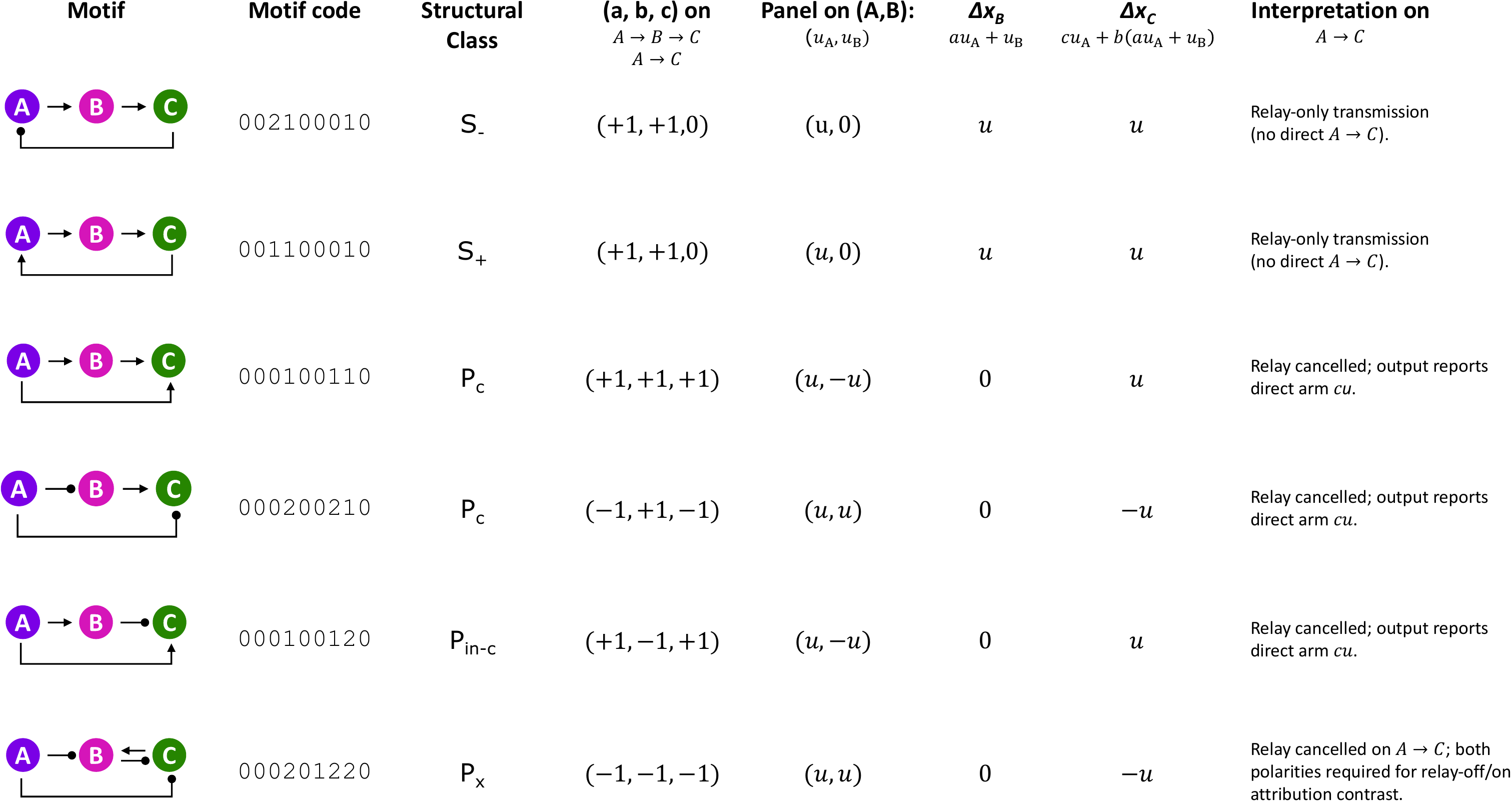}
\captionof{figure}{%
    \textbf{Relay-cancellation panel design: worked examples across structural classes.}
    Worked examples of the relay-cancellation derivation across structural classes,
    illustrating all distinct relay-cancellation polarity cases.
    Each row reports: motif barcode; structural class; signed edge indicators $(a, b, c)$
    for the parallel ordered pair $A\to C$, where $a$ (sign of first relay leg $A\to B$),
    $b$ (sign of second relay leg $B\to C$), and $c$ (sign of direct edge $A\to C$;
    $c=0$ for class $S$ motifs, which carry no direct $A\to C$ edge);
    the minimal co-perturbation panel $(u_A, u_B)$ applied to nodes $A$ and $B$; the
    relay-node response $\Delta x_B = a\,u_A + u_B$; the target response
    $\Delta x_C = c\,u_A + b(a\,u_A + u_B)$; and the mechanistic interpretation of the
    result at node $C$.
    For class $S$ motifs (rows~1--2; Rule~1), single-component perturbations produce
    relay-only transmission at $C$, with no direct arm ($c=0$).
    For class $P_c$ and $P_\mathrm{in\text{-}c}$ motifs (rows~3--5; Rule~2), the
    co-perturbation polarity is matched to the first relay leg sign: opposite-sign
    ($u_B = -u_A$) when $a=+1$ (rows~3 and~5), same-sign ($u_B = +u_A$) when $a=-1$
    (row~4), enforcing $\Delta x_B = 0$ and isolating the direct arm $c\,u_A$ at node $C$.
    For the class $P_x$ motif (row~6; Rule~3), the same-sign panel cancels the relay on
    $A\to C$ ($a=-1$); both co-perturbation polarities are required to yield the relay-off
    and relay-on attribution contrast across the mixed coherence structure.
    In the Motif column, arrows denote activating edges, filled circles denote inhibiting
    edges, and nodes are labeled A, B, C (equivalently, nodes 1, 2, 3 in the main text).
  }
\label{fig:3}
\vspace{1em}

\clearpage
\noindent\includegraphics[width=\textwidth]{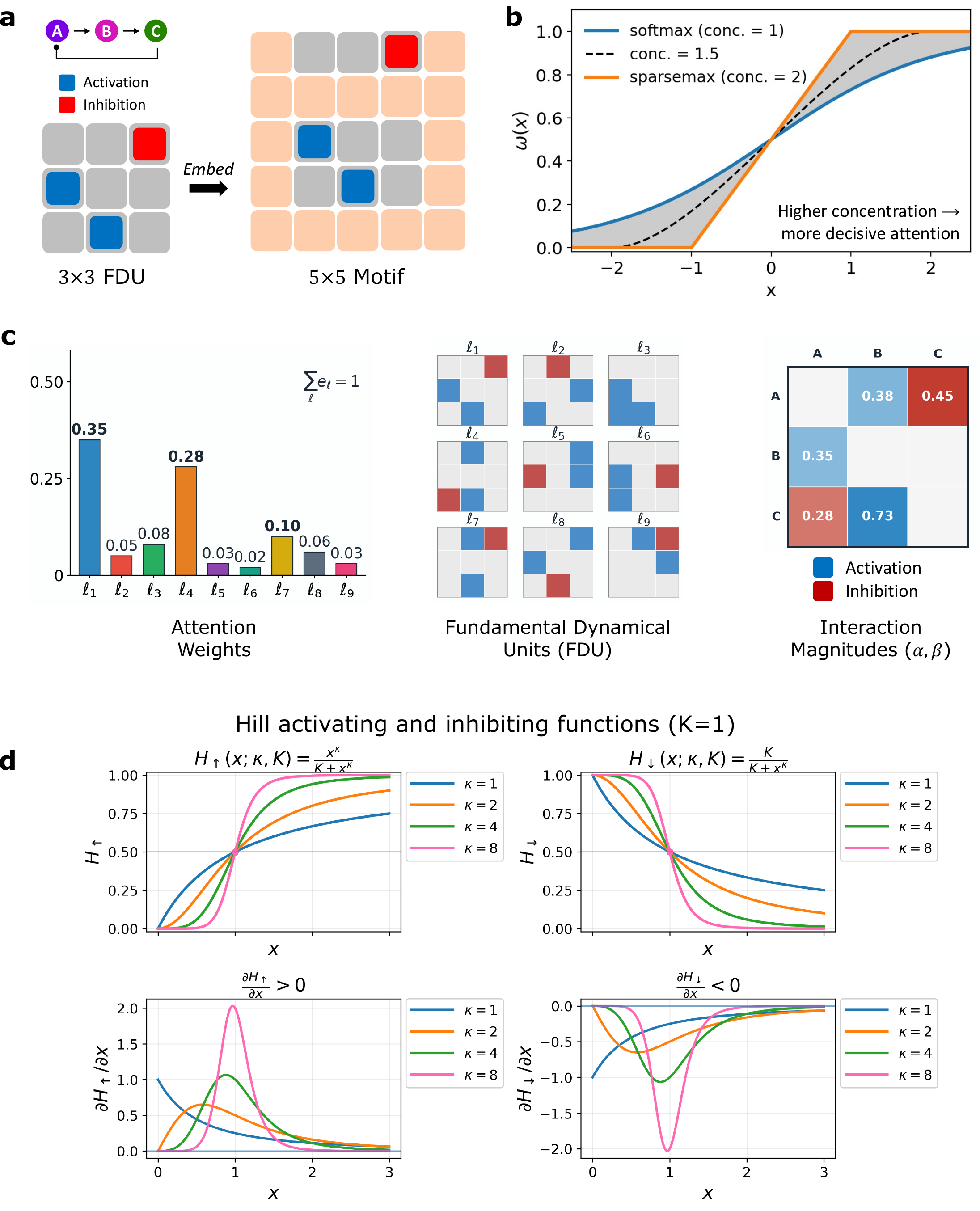}
\captionof{figure}{%
    \textbf{FDU-regularized parameterization: embedding, attention mechanism, and signed
    channel dynamics.}
    \textbf{(a)}~FDU embedding into the $N$-node setting.
    A signed three-node primitive ($3\times3$ FDU, left) is embedded onto node triplets
    within an $N$-node system to yield an $N\times N$ binary mask pair
    $(M_\uparrow^{(\ell)}, M_\downarrow^{(\ell)})$ (illustrated for $N=5$, right),
    specifying which directed edges are activating versus inhibiting under motif $\ell$.
    \textbf{(b)}~Entmax attention sparsity.
    Learnable logits $z\in\mathbb{R}^L$ are mapped to simplex-constrained attention weights
    $e = \omega(z)$ via the entmax normalization.
    Increasing the entmax concentration produces progressively more decisive (sparse)
    attention, concentrating weight on a small number of dominant FDU primitives; the
    softmax and sparsemax limits are shown for reference.
    \textbf{(c)}~The three independent components of the FDU-regularized parameterization
    (all values schematic and for illustration purposes only).
    Left: sparse entmax attention weights $e_\ell$ over the FDU bank; three dominant motifs
    (here $\ell_1$, $\ell_4$, $\ell_7$) carry the majority of attention mass.
    Center: the nine FDU motifs in the bank, each represented as a $3\times3$ binary mask
    pair ($M_\uparrow^{(\ell)}, M_\downarrow^{(\ell)}$); blue cells indicate activating edges
    and red cells indicate inhibiting edges, each at uniform full intensity.
    Right: the learned interaction magnitude matrices $\alpha$ (activating, blue) and
    $\beta$ (inhibiting, red); cell intensity reflects magnitude, encoding interaction
    strength independently of the FDU-derived structural topology.
    \textbf{(d)}~Hill-type signed channel dynamics.
    Left column: activating Hill function $H_\uparrow(x;\kappa,K)$ and its derivative
    $\partial H_\uparrow/\partial x > 0$ for cooperativity coefficients $\kappa=1,2,4,8$
    ($K=1$).
    Right column: inhibiting Hill function $H_\downarrow(x;\kappa,K)$ and its derivative
    $\partial H_\downarrow/\partial x < 0$.
    The sign-definite derivatives establish that edges with $\alpha_{ij}^\mathrm{eff}>0$
    induce small-signal activation ($\partial F_i/\partial x_j > 0$) and edges with
    $\beta_{ij}^\mathrm{eff}>0$ induce small-signal inhibition
    ($\partial F_i/\partial x_j < 0$), preserving a direct correspondence between learned
    dynamics and the signed motif representation.
  }
\label{fig:4}
\vspace{1em}

\clearpage
\noindent\includegraphics[width=\textwidth]{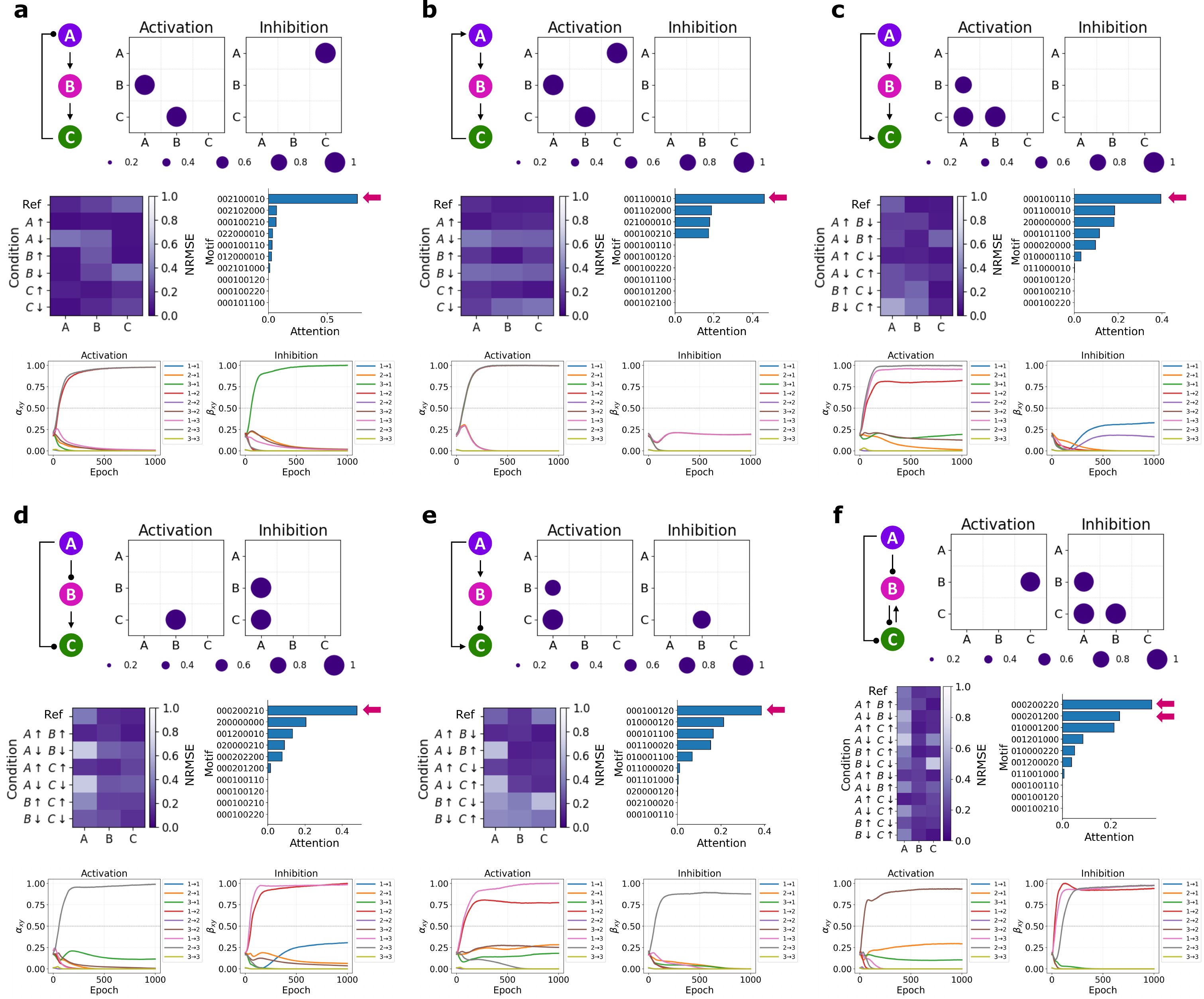}
\captionof{figure}{%
    \textbf{FDU-regularized physics-informed inference recovers signed interaction
    structure, motif identity, and dynamics across six representative motifs spanning all
    five structural classes.}
    Each panel \textbf{(a--f)} presents results for one representative motif in three rows.
    The same six motifs appear in \textbf{(Fig.~\ref{fig:3})}.
    Top row: ground-truth connectivity diagram (left; arrows denote activating edges, filled
    circles denote inhibiting edges) and inferred normalised effective interaction strengths
    as dotplots (right), with activating ($\alpha$) and inhibiting ($\beta$) channels shown
    separately; rows (y-axis) index target nodes, columns (x-axis) index source nodes, and
    dot size encodes normalised effective strength.
    Interactions at or above \mbox{50\%} of the globally normalised maximum are displayed.
    Second row: range-normalised root mean square error heatmap (rows: perturbation
    conditions; columns: state components A, B, C), evaluated on held-out time points
    (left), and FDU attention distribution over the motif bank (right), with attention
    weight on the x-axis and motif barcodes on the y-axis; the pink arrow marks the
    generating motif.
    Third row: element-wise activation ($\alpha_{xy}$, left) and inhibiting ($\beta_{xy}$,
    right) coefficient trajectories, with training epoch on the x-axis and normalised
    coefficient value on the y-axis; legend entries denote source$\to$target pairs.
    \textbf{(a)}~\texttt{002100010} (class $S_-$).
    \textbf{(b)}~\texttt{001100010} (class $S_+$).
    \textbf{(c)}~\texttt{000100110} (class $P_c$).
    \textbf{(d)}~\texttt{000200210} (class $P_c$).
    \textbf{(e)}~\texttt{000100120} (class $P_\mathrm{in\text{-}c}$).
    \textbf{(f)}~\texttt{000201220} (class $P_x$): attention distributes across two dominant
    FDUs (\texttt{000201200} and \texttt{000200220}), consistent with the unique minimal
    two-FDU decomposition for this motif.
  }
\label{fig:5}
\vspace{1em}

\clearpage
\noindent\includegraphics[width=\textwidth]{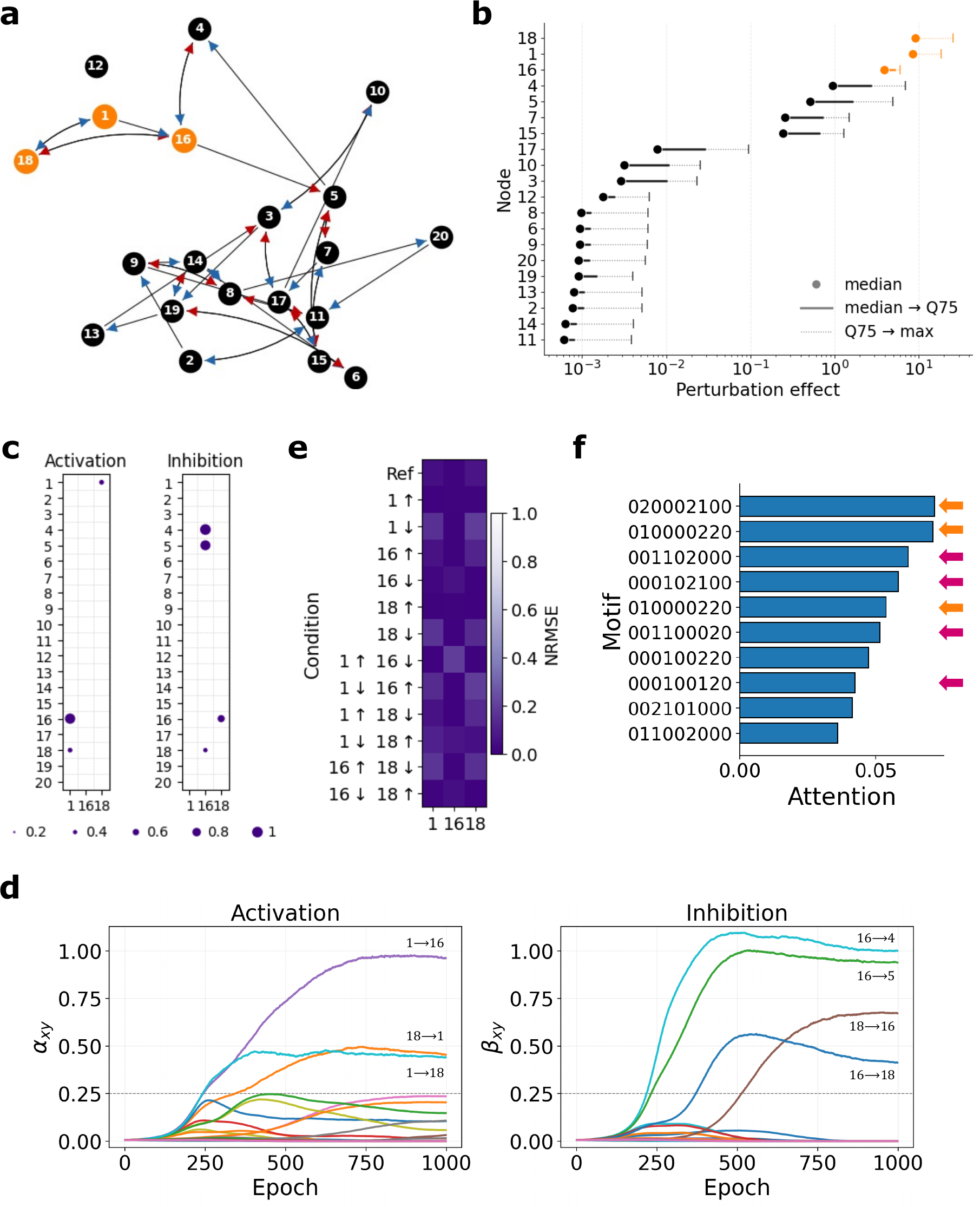}
\captionof{figure}{%
    \textbf{FDU-regularized structural inference recovers signed interaction structure and
    motif identity from a network-embedded triad under open-system confounding.}
    \textbf{(a)}~Ground-truth 20-node directed regulatory network comprising 43 directed
    signed edges across 10 overlapping signed triad subnetworks.
    Target nodes 1, 16, and 18 (orange) form a class $P_\mathrm{in\text{-}c}$ triad.
    Blue arrows: activating edges; red arrows: inhibiting edges.
    \textbf{(b)}~Per-node perturbation effect profile across all 20 nodes, with perturbation
    effect magnitude on the x-axis (log scale; measured as median
    standard-deviation-normalised root mean square error relative to the unperturbed
    reference across all non-reference conditions) and nodes sorted by effect magnitude on
    the y-axis.
    Points indicate median perturbation effect magnitude across all non-reference
    conditions; bars extend from median to Q75 (solid) and Q75 to maximum (dashed).
    Target nodes (orange) are sorted to the top.
    \textbf{(c)}~Inferred normalised effective interaction strengths as dotplots for the
    three target source columns (nodes 1, 16, 18), with activating ($\alpha$) and
    inhibiting ($\beta$) channels shown separately; rows (y-axis) index all 20 nodes, and
    dot size encodes normalised effective strength.
    Interactions at or above \mbox{25\%} of the globally normalised maximum are displayed.
    \textbf{(d)}~Element-wise activation ($\alpha_{xy}$, left) and inhibiting ($\beta_{xy}$,
    right) coefficient trajectories for the three target source columns, with training epoch
    on the x-axis and normalised coefficient value on the y-axis.
    Labeled curves identify the seven directed edges from target nodes 1, 16, and 18.
    \textbf{(e)}~Range-normalised root mean square error heatmap (rows: 13 perturbation
    conditions; columns: 3 target nodes), evaluated on held-out time points.
    \textbf{(f)}~Top 10 FDU attention weights, with attention weight on the x-axis and
    motif barcodes on the y-axis.
    Pink arrows mark the FDU components of both minimal decompositions of the target
    $P_\mathrm{in\text{-}c}$ triad.
    Orange arrows mark FDUs from the surrounding regulatory environment (ranks 1, 2,
    and~5), all involving nodes 4 and 5.
    The barcode \texttt{010000220} appears at ranks 2 and~5, reflecting two distinct
    node-set embeddings of the same FDU type, spanning nodes $\{4,5,7\}$ and $\{4,5,15\}$
    respectively; both carry confirmed edges from the full network.
  }
\label{fig:6}
\vspace{1em}

\clearpage
\begin{center}
\captionof{table}{%
    \textbf{Minimal perturbation-panel assignments across permutation-invariant structural
    classes of signed triads.}
    Counts over all 512 pairwise-connected signed triads.
    Each triad is assigned to the minimal perturbation-panel category sufficient to cancel
    relayed contributions on all parallel ordered pairs and to support attribution across
    all coherence configurations present.
    Column $(u, 0)$: single-node perturbations suffice (class $S$; no parallel ordered
    pairs) (Rule~1).
    Columns $(u, -u)$ and $(u, u)$: relay cancellation requires one co-perturbation
    polarity, determined by the first relay leg sign (Rule~2).
    Column $(u, -u\,\&\,u)$: both co-perturbation polarities, $(u, -u)$ and $(u, u)$, are
    required in the panel, arising from conflicting relay-cancellation requirements across
    parallel ordered pairs (Rule~2), or from mixed coherence structure in class $P_x$
    (Rule~3).
  }
\label{tab:panel_assignments}
\begin{tabular}{lccccc}
    \toprule
    \textbf{Class} & $(u,0)$ & $(u,-u)$ & $(u,u)$ & $(u,-u\,\&\,u)$ & \textbf{Total} \\
    \midrule
    $S_-$                    & 8 & 0  & 0  & 0   & \textbf{8}   \\
    $S_+$                    & 8 & 0  & 0  & 0   & \textbf{8}   \\
    $P_c$                    & 0 & 52 & 51 & 21  & \textbf{124} \\
    $P_\mathrm{in\text{-}c}$ & 0 & 51 & 52 & 21  & \textbf{124} \\
    $P_x$                    & 0 & 0  & 0  & 248 & \textbf{248} \\
    \bottomrule
  \end{tabular}
\end{center}
\vspace{1em}

\clearpage
\subsection{Description of Additional Supplementary Files}

\phantomsection\label{supp:data}
\noindent\textbf{Supplementary Material.}
Complete catalog of all 512 pairwise-connected signed three-node triads with structural
classifications, perturbation panel assignments, and FDU decompositions.
Each row corresponds to one admissible signed triad.
Columns: \textit{Triad ID}, the 9-character ternary motif code obtained by flattening the
$T\in\{0,1,2\}^{3\times3}$ encoding row-wise (0: no edge; 1: activating; 2: inhibiting);
\textit{Category}, the permutation-invariant structural class
($P_\mathrm{coherent}$, $P_\mathrm{incoherent}$, $P_\mathrm{mixed}$,
$S_\mathrm{positive}$, $S_\mathrm{negative}$;
corresponding to $P_c$, $P_\mathrm{in\text{-}c}$, $P_x$, $S_+$, $S_-$
in the main text);
\textit{Panel}, the minimal perturbation panel assignment prescribed by the structural class
($(u,0)$: single-component; $(u,-u)$: opposite-sign; $(u,u)$: same-sign;
$(u,-u\,\&\,u)$: both polarities);
\textit{Number of mutual pairs}, the number of unordered node pairs carrying edges in both
directions; \textit{Number of edges}, the total number of directed signed edges;
\textit{Number of compatible FDUs}, the number of tournament FDUs compatible with the
triad under sign-consistent superposition; \textit{FDU Decompositions}, the minimal FDU
decomposition(s) as pipe-separated barcode pairs, with multiple decompositions separated
by commas.

\vspace{1em}
\phantomsection\label{supp:fig1}\label{supp:fig2}\label{supp:fig3}\label{supp:table1}\label{supp:notation}
\noindent\textbf{Supplementary Information.}
Contains \textbf{Supplementary Fig.~1}, \textbf{Supplementary Fig.~2},
\textbf{Supplementary Fig.~3}, \textbf{Supplementary Table~1}, and \textbf{Supplementary Note}.

\end{document}